\documentclass[a4paper,fleqn]{cas-sc}

\usepackage[authoryear,longnamesfirst]{natbib}
\usepackage{graphicx}
\usepackage{lscape}
\usepackage{caption}
\usepackage{enumitem}
\usepackage{booktabs}
\usepackage{gensymb}
\usepackage{subfig}
\usepackage{bbm}
\usepackage{float}
\usepackage{adjustbox}
\usepackage{amsmath}

\newcommand{\mypar}[1]{\vspace{0.15em}
\textbf{#1}~}

\usepackage{xcolor}   
\usepackage{colortbl}

\newcommand{\fImg}{\mathbf{I}}

\newcommand{\xx}{\mathbf{x}}

\newcommand{\data}{\mathcal{D}}

\usepackage[ruled,vlined]{algorithm2e}

\definecolor{commentcolor}{RGB}{110,154,155}   
\newcommand{\PyComment}[1]{\ttfamily\textcolor{commentcolor}{\# #1}}  
\newcommand{\PyCode}[1]{\ttfamily\textcolor{black}{#1}}

\def\tsc#1{\csdef{#1}{\textsc{\lowercase{#1}}\xspace}}
\tsc{WGM}
\tsc{QE}
\tsc{EP}
\tsc{PMS}
\tsc{BEC}
\tsc{DE}

\begin{document}
\let\WriteBookmarks\relax
\def\floatpagepagefraction{1}
\def\textpagefraction{.001}

\shorttitle{Domain Adaptation Techniques for Natural and Medical Image Classification}

\shortauthors{Ahmad Chaddad et~al.}

\title [mode = title]{Domain Adaptation Techniques for Natural and Medical Image Classification}                      



%
\author[add1,add2]{Ahmad Chaddad \corref{cor1}}[orcid=0000-0003-3402-9576]
\ead{ahmadchaddad@guet.edu.cn}
\author[add1]{Yihang Wu}
\author[add3,add4]{Reem Kateb}
\author[add2]{Christian Desrosiers}

\address[add1]{School of Artificial Intelligence, Guilin University of Electronic Technology, Guilin, China, 541004}
\address[add2]{The Laboratory for Imagery, Vision and Artificial Intelligence, Ecole de Technologie Superieure, Montreal, Canada, H3C 1K3}
\address[add3]{College of Computer Science and Engineering, Taibah University, Madinah, Saudi Arabia, 42353}
\address[add4]{College of Computer Science and Engineering, Jeddah University, Jeddah, Saudi Arabia, 23445}

\fntext[fn1]{Ahmad Chaddad and Yihang Wu are equally contributed.} 


\begin{abstract}
Domain adaptation (DA) techniques have the potential in machine learning to alleviate distribution differences between training and test sets by leveraging information from source domains. In image classification, most advances in DA have been made using natural images rather than medical data, which are harder to work with. Moreover, even for natural images, the use of mainstream datasets can lead to performance bias. {With the aim of better understanding the benefits of DA for both natural and medical images, this study performs 557 simulation studies using seven widely-used DA techniques for image classification in five natural and eight medical datasets that cover various scenarios, such as out-of-distribution, dynamic data streams, and limited training samples.} Our experiments yield detailed results and insightful observations highlighting the performance and medical applicability of these techniques. Notably, our results have shown the outstanding performance of the Deep Subdomain Adaptation Network (DSAN) algorithm. This algorithm achieved feasible classification accuracy (91.2\%) in the COVID-19 dataset using Resnet50 and showed an important accuracy improvement in the dynamic data stream DA scenario (+6.7\%) compared to the baseline. Our results also demonstrate that DSAN exhibits remarkable level of explainability when evaluated on COVID-19 and skin cancer datasets. These results contribute to the understanding of DA techniques and offer valuable insight into the effective adaptation of models to medical data. 
\end{abstract}




\maketitle

\section{Introduction}
Rapid progress in the field of artificial intelligence (AI) has resulted in the development of several powerful techniques, the most prominent of which is deep learning \cite{van2023unpaired}. {Deep learning techniques have shown impressive performance in areas such as computer vision and medical imaging \cite{10835760}, however, they often require a significant number of well-annotated training data to achieve high levels of performance.} Obtaining such data can be a difficult undertaking, especially in the healthcare field \cite{van2023unpaired}.

\begin{figure}[htp]
    \includegraphics[width = 0.98 \textwidth]{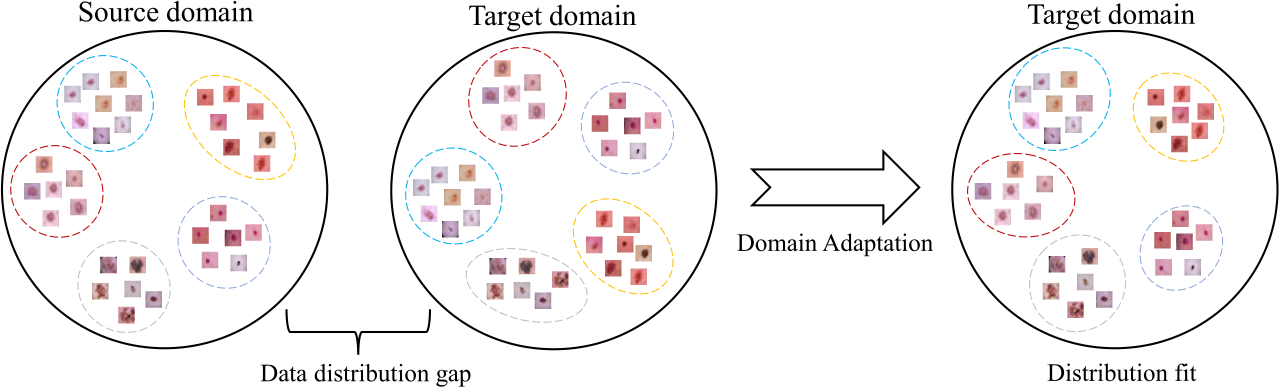}
    \caption{Examples of domain adaptation (DA). (\textbf{Left}): There exists a large gap between source and target domain data distributions. (\textbf{Right}): After applying DA, the data distribution discrepancies decreased, and the target domain data distribution is more fit with the source data. The color circles represent different classes.}
    \label{F:DA_Intro}
\end{figure}

A well-known strategy to alleviate this problem, known as domain adaptation (DA), is to find another data source for the same task, but with different characteristics, and then adapt a model trained in this source to also work on the target \cite{10835760}. More generally, DA refers to learning scenarios where the source and target tasks are identical but the data distributions of the source and target domains differ. The goal of DA is to reduce differences in data distributions between different domains. Figure \ref{F:DA_Intro} illustrates how DA is used to reduce these discrepancies. The model trained on the source domain is better suited for implementation in the target domain. {We note that DA can be categorized into several types, such as traditional DA (e.g., feature-based methods) and deep DA (e.g., discrepancy-based methods) \cite{peng2022domain}. However, these classifications may vary depending on the new techniques introduced in this context.}

Numerous DA methods have been proposed by researchers, who have subsequently applied them to a diverse range of problems. For traditional DA, for example, in \cite{sun2017correlation}, they argued that DA could be achieved by aligning the covariance of the source and target data, proposing a correlation alignment (Coral) and showing feasible performance in the Office dataset. {Similar to maximum mean discrepancy (MMD), simply applying correlation alignment on all classes leads to poor adaptation results for rare class samples. In deep DA, the categories can be divided into four main parts: deep MMD-based (i.e., measure and minimize the distribution differences between source and target domains in a kernel-induced feature space), deep correlation-based (i.e., reduce the correlation between source and target domain features to learn a shared representation where the domains are de-correlated, thereby improving transferability), adversarial techniques (i.e., use a discriminator network to distinguish between source and target domain features, and the generator network is trained together with the discriminator, thus aligning the distributions in an adversarial manner), and others (e.g., norm maximization).} {In deep MMD-based methods, for example, in \cite{zhu2020deep}, they further extended multi-kernel MMD to the subdomain MMD, using class information to align features in the source and target domains.} {Their approach addresses the mis-aligment introduced by global MMD, thereby improves the adaptation effects to rare class samples.} Furthermore, in \cite{ge2023unsupervised}, the authors proposed using conditional MMD (CMMD) with mutual information (MI) extracted from unlabeled target data for unsupervised DA. The experimental results in the Office31 and OfficeHome datasets demonstrate the usefulness of their methodology. {But, it requires higher computation time due to the calculation of the kernel matrix. In addition, performance improvements on many easily confused classes are marginal.} 


For deep correlation-based, in \cite{sun2016deep}, they extended Coral to deep Coral, indicating feasible performance compared to MMD-based methods. {However, it faces the same challenge of traditional Coral technique (i.e., mis-alignment for rare class samples).} {Furthermore, in \cite{10856559}, they proposed a novel heterogeneous DA method that learns classification-correlative and discriminative feature representations by maximizing class-specific feature correlations and interclass distribution distances while minimizing marginal and conditional distribution divergences. They further incorporate a selective pseudo-labeling procedure to improve performance. However, it fails to generalize to noisy data and limits its potential where the source domain contains large noise. In \cite{10820825}, they propose a novel weakly correlated multimodal DA method that acquires modality-independent and category-related knowledge from the source domain by extracting modality-invariant and domain-invariant features, using a source-specific classifier with pseudo-labels and a target-specific classifier with highly reliable pseudo-labels for effective knowledge transfer. However, the performance of their method is degraded when the modality information is limited.}

Despite the adaptation techniques such as MMD, correlation alignment mentioned above, in \cite{ganin2016domain}, they introduced an adversarial domain classifier to perform DA, highlighting the potential of adversarial networks. {However, it does not guarantee alignment of conditional distributions, thus fails to capture class-specific alignments between the source and target domains.} In \cite{DALN}, inspired by Batch Nuclear Maximization (BNM) \cite{BNM}, they proposed to reuse the task specific classifier as discriminator, highlighting the potential of discriminator free adversarial DA. But, it introduces the risk of overfitting to the source domain as it reuses the same classifier to classify the source domain samples and align the source domain feature with target domain. {In \cite{wu2025faa}, they explored the usefulness of foundation models (FMs) such as vision language models (VLMs) to enrich the image feature qualities, and use a domain classifier to distinguish the source / target domain for adversarial DA in a federated learning (FL) framework. With the rich features extracted by contrastive language image pre-training (CLIP), their method outperforms other methods with large margins. Furthermore, their approach requires less computational and communication overhead, demonstrating the potential of FMs in FL-based DA.} 

For other methods, in \cite{BNM}, they argued that target features nuclear norm can be used to perform DA. Their method demonstrate favorable accuracy in Office31, OfficeHome datasets. However, the BNM needs large computational costs. {In \cite{zhang2025domain}, they proposed a novel Domain-guided Conditional Diffusion Model to address the limitations of unsupervised DA (UDA) by generating high-quality target domain samples, which are guided by class information and a domain classifier, thereby improving UDA performance on Office31 and OfficeHome datasets. However, when the source domain has limited samples, the diffusion model does not provide high-quality generated images. This leads to lower performance compared to non-diffusion-based approaches. Furthermore, in \cite{cai2025multi}, they proposed a new DA framework that incorporated an adaptive multimodal diffusion network and a deep causal inference reweighting mechanism. This framework addresses the limitations of GAN-based DA methods and outperforms existing methods on various real-world DA datasets. However, their diffusion network introduced considerable computational overhead. In addition, in \cite{10935814}, they proposed a novel DA framework for remote image sensing with an adaptation diffusion model distillation module and a consistent causal intervention (CCI) module. Using the diffusion model, their method shows a better feature alignment effect. However, it requires four times more computation time than ResNet50. In \cite{liu2024textadapter}, they explored self-supervised DA with the generative model. They use data augmentation to provide multiple inputs, while leveraging contrastive learning and consistency regularization to improve performance. Moreover, in \cite{wang2025samda}, they explored FMs (e.g., segment anything model) for medical image segmentation with DA. They minimize the euclidean distance between the gram matrices of the source and target data to enhance the feature representation ability of the model. The experimental results show that it yields a higher Dice score compared to the baselines.} {In \cite{yu2024FFTAT}, they introduced an innovative FM-based approach (i.e., transformer model) for unsupervised domain adaptation tasks. This involved employing a patch discriminator to build a transferability matrix which directs self-attention towards transferable patches. In addition, they incorporated a feature fusion technique to enhance generalization by allowing embeddings to integrate information from all domains. Their method achieved an average accuracy of 51.9\% on DomainNet, presenting a feasible solution for large-scale DA tasks using transformer models. In \cite{jang2024robust}, they designed a negative-view regularization term to optimize the feature representation of FMs, such as vision transformers (ViT), for DA tasks. Experimental results show that the proposed approach provides more clear decision boundaries than without a regularization term. Furthermore, their method demonstrates reasonable consistency between the attention map and the prediction, highlighting its reliability. Furthermore, in \cite{yu2025open}, they designed a fine-tuning framework for FMs with open-set DA. They used a pre-trained CLIP model to guide a convolutional neural network (CNN) in predicting images. The experimental results show that the proposed method outperforms traditional DA techniques, suggesting that FMs can lead to better adaptation abilities.}

\begin{figure}[htp]
    \includegraphics[width = 0.98 \textwidth]{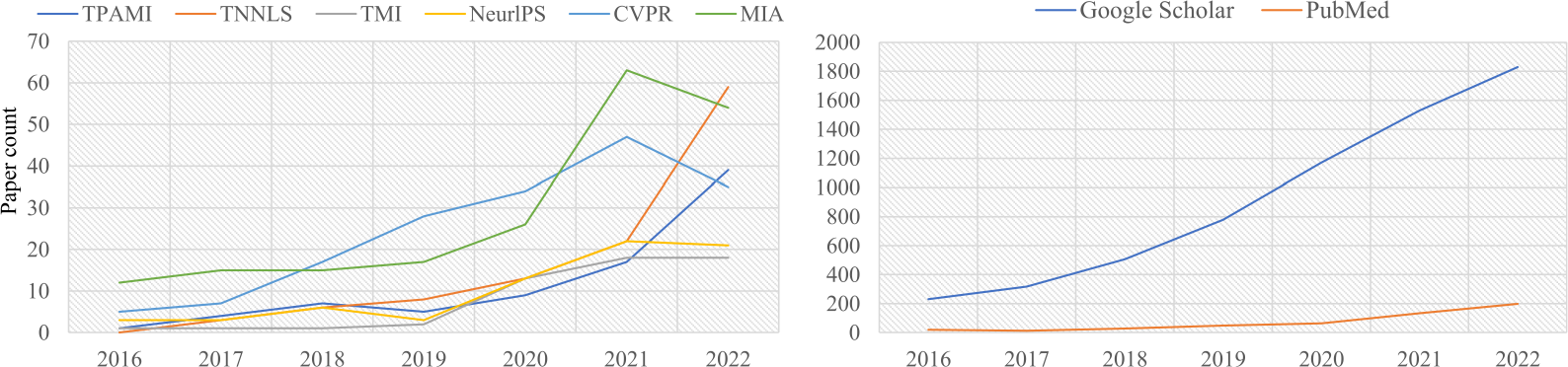}
    \caption{\textbf{Left}: Example of paper number related to domain adaptation each year in top venues journals and conferences). \textbf{Right}: Search results on Google Scholar and PubMed database. Specifically, we employ the following search method: for CVPR and NeurIPS, we only considered papers that have ``domain adaptation'' in their titles. For other journals (conferences), we recorded papers that have ``domain adaptation'' either in their titles or in the keywords. The paper titled ``Domain Adaptation'' was included in the count for both Google Scholar and PubMed.}
    \label{F:SearchResult}
\end{figure}

Figure \ref{F:SearchResult} summarizes the search results for DA-related articles that have been published in leading journals or conferences, with the results obtained from the Google Scholar and PubMed databases for the period between 2016 and 2022. As shown, this field has been the focus of a remarkable number of papers in recent years.

Despite the numerous advances in DA, it is important to note that most of the results are still derived from standard datasets such as Office31 \cite{Office31} and others. However, optimizing the performance of DA algorithms for mainstream datasets can potentially result in substantial bias in the results. If these algorithms are tested on non-standard datasets, such as more realistic hospital data or newly proposed natural datasets, their performance may not be as impressive. In other words, their ability to generalize may be reduced when applied to different datasets. Moreover, there is a lack of comprehensive studies to compare popular DA algorithms for classification. In \cite{zhu2020deep}, the authors used classification accuracy, visualization of t-SNE plots, and $\mathcal{A}$-distance to demonstrate the usefulness of their approach. On the contrary, the study that introduced Deep Coral \cite{sun2016deep} only considered accuracy as a performance measure, which may not be sufficient to reflect the effectiveness of its method in addressing data distribution problems. {Selecting the most appropriate DA technique for a specific task presents a challenge due to the lack of comprehensive analysis and comparison.}  With such questions, we selected several DA algorithms commonly used for natural image classification, such as Deep Coral \cite{sun2016deep}, Deep Adversarial Neural Network (DANN) \cite{ganin2016domain}, Deep Subdomain Adaptation Network (DSAN) \cite{zhu2020deep}, Batch Nuclear Norm Maximization (BNM) \cite{BNM}, Discriminator Free Domain Adaptation (DALN) \cite{Chen_2022_CVPR} and Deep Conditional Adaptation Network (DCAN) \cite{ge2023unsupervised}, and tested them on natural and medical images. Our work extends the standard definition of DA considering it as a technique to improve the performance of a classifier in a target domain by training it with the source data. Whether the dataset contains natural or medical images, the considered DA models involve a training set (source domain) and a testing set (target) with significant data distributions. {Unlike the previous state-of-the-art related to DA \cite{farahani2021brief,fang2024source}, the novelty of this study lies in a comprehensive simulation study (557 experiments) using seven common and new DA methods with 13 public datasets related to natural and medical images. Specifically, this study covers different data scales (i.e., the number of samples $\in (100, 400000)$), task complexities (i.e., the number of classes $\in (2, 200)$), and different modalities (i.e., Computed Tomography (CT), Magnetic Resonance Imaging (MRI) and Dermoscope images).} The main contributions of this paper are as follows.
\begin{itemize}
    \item {We simulate seven commonly-used and novel DA algorithms based on theoretical distinctions (i.e., correlation-, MMD-, adversarial- and other based techniques) for image classification using natural (five datasets) and medical (eight datasets) images.}
    \item {We evaluate the performance of these DA models in a variety of scenarios, such as out-of-distribution, limited training samples, and dynamic data-stream settings.} 
    \item {We investigate the integration of explainable artificial intelligence (XAI) techniques within the DA classifiers to enhance their understandability. The results suggest that the use of DA does not guarantee reasonable visual explanations.}
    \item {We discuss our experimental results from multiple perspectives (e.g., data quality, texture shifts, data privacy) and highlight the potential challenges and future directions of DA. }
\end{itemize}

The remaining sections of this paper are organized as follows. Section \ref{S:2} provides a brief overview of the six common/novel DA techniques. Section \ref{S:3} presents the experimental results obtained using these techniques, together with a comprehensive analysis. {Section \ref{S:4} illustrates the experimental results of these methodologies in various situations and settings.} In Section \ref{S:5}, we discuss important insights derived from our results and present future trends in DA. {Lastly, in Section \ref{S:6}, we provide our
concluding remarks.} 

\section{Background on common and recent algorithms}\label{S:2}
As mentioned previously, DA methods take advantage of machine learning techniques that improve model performance when faced with a distribution shift between training data and test data. Suppose $f$ is a classifier trained in the source domain, the goal of DA is to minimize the classification loss in the target domain using the classifier $f$. Despite the advancements of many DA techniques, however, the key concept remains the same: align the data distribution in the source and target domain in the same feature space, thus increasing the transferability of the model trained in the source. {To explore the development of DA approaches, we focus on the six commonly used and one recent DA algorithms as follows.} 

\subsection{Deep correlation based methods}

\textbf{Deep Coral:} The authors of \cite{sun2016deep} introduce a differentiable loss function in their network to align the features of sources and target domains. The Deep Coral approach extends the standard Coral method by learning non-linear transformations that are more powerful and compatible with deep neural networks. The goal is to minimize the difference in covariance between the features learned from different domains. Formally, their optimization objective can be written as
\begin{equation}
    \ell_{Coral} \, = \, \frac{1}{4 d^2}\left\|C_S-C_T\right\|_F^2
\end{equation}
where $d$ represents the dimension of the neural network's output. $C_S$ and $C_T$ represent the covariance of the source and target data, respectively.

{Thus, $C_S$ and $C_T$ can be defined as:}
\begin{equation}
\begin{aligned}
&C_S=\frac{1}{n_S-1}\left(\mathcal{D}_S^{T} \mathcal{D}_S-\frac{1}{n_S}\left(\mathbf{1}^{T} \mathcal{D}_S\right)^{T}\left(\mathbf{1}^{T} \mathcal{D}_S\right)\right)
\end{aligned}
\end{equation}
\begin{equation}
\begin{aligned}
&C_T=\frac{1}{n_T-1}\left(\mathcal{D}_T^{T} \mathcal{D}_T-\frac{1}{n_T}\left(\mathbf{1}^{T} \mathcal{D}_T\right)^{T}\left(\mathbf{1}^{T} \mathcal{D}_T\right)\right)
\end{aligned}
\end{equation}
{where $\mathcal{D}_S$ and $\mathcal{D}_T$ indicate the source and target data, respectively. $n_S$ and $n_T$ represent the number of samples in source and target domain, respectively.}

\subsection{Deep MMD based approaches}

\textbf{Deep subdomain adaptation network (DSAN): } The concept of ``sub-domain'' was introduced in \cite{zhu2020deep}. Samples within a domain can be divided into sub-domains based on different conditions. One way to divide them is by using class labels, which results in sub-domains that contain the same classes. This approach is commonly used in research, where the category serves as the basis for the division. Instead of performing a global alignment, localized sub-domains are matched separately. The final goal can be represented as follows.
\begin{equation}
{\hat d_{\cal H}}(\mathcal{D}_s,\mathcal{D}_t) \, = \, \frac{1}{C}\sum\limits_{c = 1}^C {\Bigg\| {\sum\limits_{x_i^s \in \mathcal{D}_s} {\omega _i^{sc}} \phi \left( {x_i^s} \right) - \sum\limits_{x_j^t \in \mathcal{D}_t} {\omega _j^{tc}} \phi \left( {x_j^t} \right)} \Bigg\|_{\cal H}^2} 
\label{Eq:8}
\end{equation}
Here, $\omega^c$ is defined according to the number of samples in each category, C is the number of classes. {$\phi(\cdot)$ is a mapping function.} $x^s$ and $x^t$ are the occurrences in datasets $D^s$ and $D^t$, correspondingly. The variables $D^s$ and $D^t$ refer to the source and target domains, respectively. {In practice, Eq. \ref{Eq:8} can be reformulated using kernel function to calculate the estimation of this discrepancy:}
\begin{equation}
    {\hat d_H}({D_s},{D_t}){\mkern 1mu}  = {\mkern 1mu} \frac{1}{C}\sum\limits_{c = 1}^C {\left[ {\sum\limits_{i = 1}^{{n_s}} {\sum\limits_{j = 1}^{{n_s}} {\omega _i^{sc}\omega _j^{sc}k(x_i^s,x_j^s) + \sum\limits_{i = 1}^{{n_t}} {\sum\limits_{j = 1}^{{n_t}} {\omega _i^{tc}\omega _j^{tc}k(x_i^t,x_j^t) - 2\sum\limits_{i = 1}^{{n_s}} {\sum\limits_{j = 1}^{{n_t}} {\omega _i^{sc}\omega _j^{tc}k(x_i^s,x_j^t)} } } } } } } \right]}
\end{equation}
{where $n_s$ and $n_t$ are the number of source and target samples, respectively. $k(\cdot, \cdot)$ is the kernel function.}

{\mypar{Deep Conditional Adaptation Network (DCAN):} In \cite{ge2023unsupervised}, the authors extended the work of DSAN, proposing a novel CMMD based feature adaptation technique with MI to align the source and target domain for unsupervised DA. CMMD aligns the conditional distributions of $\mathcal{D}_s$ and $\mathcal{D}_t$. However, CMMD alone may not effectively extract discriminant information from the target domain. Thus, they introduced MI to extract representative features in the target domain by maximizing MI. In practice, their CMMD loss can be reformulated as:}

\begin{equation}
    {\mathcal{L}_{CMMD}} = Tr({L_s}\tilde L_s^{ - 1}{K_s}\tilde L_s^{ - 1}) + Tr({L_t}\tilde L_t^{ - 1}{K_t}\tilde L_t^{ - 1}) - 2 \cdot Tr({L_{ts}}\tilde L_s^{ - 1}{K_{st}}\tilde L_t^{ - 1})
\end{equation}

where $Tr(\cdot)$ is the trace of the matrix, ${\mathcal{L}_s} = \Phi _s^{\rm T}{\Phi _s}$ is the kernel function, ${\Phi _s} = [\phi (y_1^s),...,\phi (y_n^s)]$. ${K_s} = \Psi _s^{\rm T}{\Psi _s}$, ${\Psi _s} = [\psi (x_1^s),...,\psi (x_n^s)]$, $\phi(\cdot)$ and $\psi(\cdot)$ are mapping functions, respectively. ${{\tilde L}_s} = {L_s} + \textbf{I}$, $\textbf{I}$ is a matrix with full ones. $\Psi_t$, $\Phi_t$, $K_t$, $L_t$ and $\hat{L}_t$ are defined similarly.

Furthermore, MI loss can be defined as:
\begin{equation}
    {\mathcal{L}_{MI}} = \frac{1}{n}\sum\limits_{i = 1}^n {H(\hat y_i^t\left| {x_i^t)} \right.}  - H[\frac{1}{n}\sum\limits_{i = 1}^n {P(\hat y_i^t\left| {x_i^t)} \right.} ]
\end{equation}
{where ${H(\hat y_i^t\left| {x_i^t)} \right.}$ denotes the conditional entropy, ${P(\hat y_i^t\left| {x_i^t)} \right.}$ is the corresponding prediction classification probability.}

\mypar{Efficient unsupervised domain adaptation (EUDA):} {In \cite{abedi2024euda}, they explored the combination of FMs such as DINOv2 (self-supervised vision transformer) with the synergistic domain alignment (SDA) loss (i.e., cross-entropy and MMD losses). Specifically, they use the pre-trained DINOv2 as a feature extractor to extract rich image features, while introducing a bottleneck layer (i.e., fully connected layers with activation and normalization layers) and a classifier head (i.e., a fully connected layer, shared with target domain) to learn task-related knowledge. Furthermore, SDA loss is used to fine-tune these layers. Their MMD loss can be represented as follows:}
\begin{equation}
{\hat d_{\cal H}}(\mathcal{D}_s,\mathcal{D}_t) \, = \, {\Bigg\| {\frac{1}{|\mathcal{D}_s|}\sum\limits_{x_i^s \in \mathcal{D}_s} {} \phi \left( {x_i^s} \right) - \frac{1}{|\mathcal{D}_t|}\sum\limits_{x_j^t \in \mathcal{D}_t} {} \phi ( {x_j^t})} \Bigg\|_{\cal H}^2} 
\end{equation}
{where $|\mathcal{D}_{s|t}|$ represents the number of samples in the source (target) domain, $\phi(\cdot)$ is a mapping function similar to DSAN.}

\subsection{Deep adversarial and other DA techniques}

\textbf{Deep adversarial neural network (DANN):} In the field of DA, the integration of generative adversarial networks (GANs) was first proposed in \cite{ganin2016domain}. This strategy involves adding a domain classifier to the network architecture, which tries to predict the domain of the input data. The main idea is to encourage the network to learn domain-invariant features by making the domain classifier perform poorly. This helps the network focus on shared patterns across domains and reduces the impact of domain-specific variations. By incorporating domain-adversarial training, the neural network becomes more robust and effective in handling data from different domains. {In each iteration, batches of $2B$ samples are generated randomly by selecting the same number of source and target samples. The adversarial DA loss is defined using cross-entropy, as follows:}
\begin{equation}\label{eq:total_loss}
\begin{aligned}
   \mathcal{L}_{adv} \, = \, - \frac{1}{2B} \sum_{j=1}^{2B} z_j\log D_i(\fImg_j)
   \, + \, (1-z_j)\log \big(1-D_i(\fImg_j)\big)
\end{aligned}
\end{equation}

{where $z_j = \mathbbm{1}(\xx_j \in \data_i)$ the domain label of example $\xx_j$. $\textbf{I}_j$ are the features extracted by $e_f$. The domain classifier $D$ will predict if a given image representation $\fImg_j$ is from a \emph{source} ($D_i(\fImg_j) =1$) or a \emph{target} domain ($D_i(\fImg_j) =0$). }

{\mypar{Discriminator-Free Adversarial Domain Adaptation (DALN):} Previous adversarial based DA techniques assume that there exists a discriminator; however, in \cite{Chen_2022_CVPR}, the authors argued that feature alignment can be achieved by reusing the task specific classifier as discriminator, demonstrating favorable accuracy in Office31 dataset. In general, they proposed a Nuclear-norm Wasserstein discrepancy (NWD) for performing discrimination. Suppose $\textbf{x}_s,\textbf{x}_t$ are the source and target domain data, respectively. Their NWD loss can be defined as follows. }
\begin{equation}
    {\mathcal{L}_{NWD}}({\textbf{x}_s},{\textbf{x}_t}) = \frac{1}{{{N_s}}}\sum\limits_{i = 1}^{{N_s}} {D(G(x_s^i} )) - \frac{1}{{{N_t}}}\sum\limits_{i = 1}^{{N_t}} {D(G(x_t^i} ))
\end{equation}
{where $D$ represents the task specific classifier, $G$ indicates the feature extractor (e.g., Resnet50).}

The final loss function can be reformulated through a min-max optimization as:
\begin{equation}
    \mathop {\min }\limits_G \mathop {\max }\limits_D \mathcal{L} = {\mathcal{L}_{CE}}({x_s},{y_s}) - \lambda {\mathcal{L}_{NWD}}({x_s},{x_t})
\end{equation}
where $\lambda$ is a hyper-parameter controlling the trade-off between these two loss terms. In practice, this min-max optimization problem is converted to standard minimization using a gradient reversal layer \cite{ganin2016domain}.

\textbf{Batch nuclear norm maximization (BNM): } BNM is a powerful technique used in the field of DA to effectively tackle distribution differences \cite{BNM}. The goal of this technique is to learn a transformation matrix that reduces the differences between the source and target domains. BNM uses the nuclear norm, which is also referred to as the trace norm, to regularize the transformation matrix and promote solutions with low rank. The nuclear norm refers to the sum of the singular values of a matrix. It is commonly used as a convex substitute for the rank function. The BNM method encourages a low-rank structure in the transformation matrix by maximizing the nuclear norm, which can effectively reduce the distribution discrepancy between the source and target domains. {We suppose $G$ is the neural network, $\textbf{x}^t$ consists of a batch of target domain data, the BNM loss can be formulated as:}
\begin{equation}
    {\mathcal{L}_{BNM}} =  - \frac{1}{B}{\left\| {G({x^t})} \right\|_*}
\end{equation}
{where $B$ is the batch size, ${\left\|  \cdot  \right\|_*}$ is the nuclear norm of a matrix.}

\section{Experiments using various datasets}\label{S:3}

\begin{table}[htp] \footnotesize
    \caption{Summary of the implementation details of each algorithm.}
    \setlength{\tabcolsep}{0.58cm}
    \begin{tabular}{ccccccccc}
    \toprule
     Alg. & LR & Epoch & BS &M & OS &WD & Iter & $\lambda$ \\ 
     \midrule
DC& $3 \times 10^{-3}$ & 30 & 16&0.9 &SGD&$5\times 10^{-4}$&200&10 \\  
     DANN&$1 \times 10^{-2}$ &30 &16&0.9&SGD&$1\times 10^{-3}$&200&1\\ 
DSAN &$1\times 10^{-2}$ &30 &16&0.9&SGD&$5\times 10^{-4}$&200&0.5 \\ 
     BNM  &$1\times 10^{-3}$ &30 &16&0.9&SGD&$5\times 10^{-4}$&200&1 \\
     DALN&$1\times 10^{-2}$&30&16&0.9&SGD&$1 \times 10^{-3}$& 200 & 0.1 \\
     DCAN&$1\times 10^{-2}$&30&16&0.9&SGD&$5 \times 10^{-4}$& 200 & 0.5 \\
     EUDA&$3\times 10^{-2}$&30&16&0.9&SGD&0& 200 & 0.3 \\
 w/o DA & $1 \times 10^{-2}$ & 30  &16 & 0.9 &SGD&$5 \times 10^{-4}$&200&0 \\
     \bottomrule
    \end{tabular}
    {LR: Learning rate; BS: Batch size; M: Momentum; OS: Optimization strategy; SGD: Stochastic gradient descent; WD: Weight decay; Iter: iteration per epoch; DC: Deep Coral; $\lambda$: domain adaptation loss weights.}
    \label{T:Details_Alg}
\end{table}

\subsection{Data pre-processing process and experimental settings}
\begin{figure*}[htp]
     \centering
    \includegraphics[width = 0.96 \textwidth]{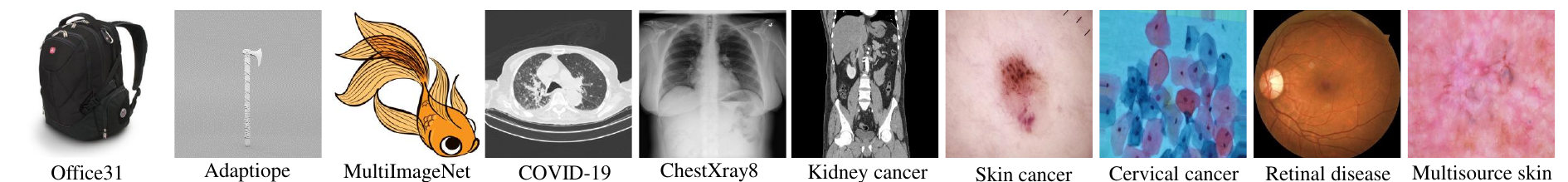}
     \caption{Example of samples from each dataset.}
    \label{F:Dataset}
\end{figure*}

Images were pre-processed before use in neural network training. Each image was first resized to dimensions of 256 $\times$ 256. Then a random cropping operation was applied to the image, resulting in a final size of 224 $\times$ 224. Furthermore, the image was randomly flipped horizontally with a probability of 0.5 and the pixel values were normalized using z-score normalization. Figure \ref{F:Dataset} shows samples from the different datasets. Our testing environment is based on Windows 11, with an Intel 11900K CPU, 128 GB RAM, and an RTX 2060 Graphic Card. 

\begin{figure*}[htp]
     \centering
    \includegraphics[width = 0.96 \textwidth]{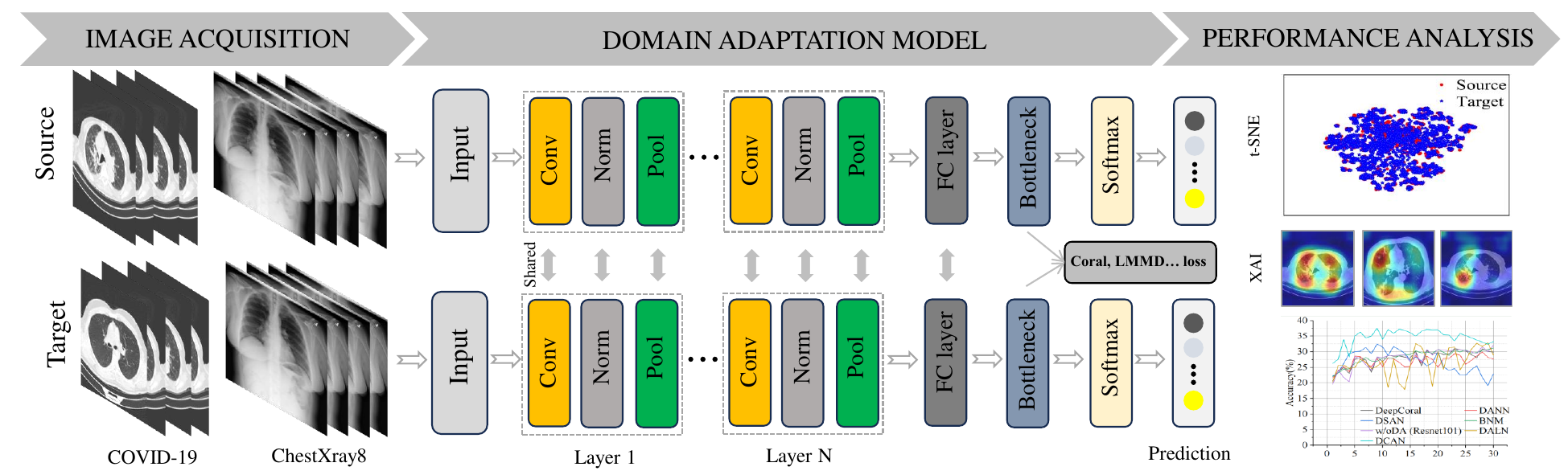}
    \caption{The simulation pipeline and its stages. The training process for backpropagation involves the computation of the domain adaptation loss between the source and target domains at the bottleneck layer, along with the cross-entropy loss on the source domain.}
    \label{F:Pipeline}
\end{figure*}

For training, we used an optimizer based on the mini-batch stochastic gradient descent (SGD) strategy \cite{bottou2010large}. The neural network's loss function, which comprises two distinct components, can be formulated as follows:
\begin{equation}\footnotesize
    loss \, = \, CrossEntropy(y_{pred}^s,y_{true}^s) \, + \, \lambda\, \mathrm{DA}\!\left( {{D_s},{D_t}} \right)
\end{equation}
In this loss, $D_s$, $D_t$ are the source and the target data. The initial component corresponds to the cross-entropy loss associated with the source domain. The second component encourages alignment between the data from both the source and the target domains. Figure \ref{F:Pipeline} shows the pipeline of our testing procedure.

\subsection{Evaluation metrics}
The methods' effectiveness was assessed by observing the change in classification accuracy before and after the use of DA. Given the number of total samples $N$, the model's prediction $y_{pred}$, ground-truth label $y_{true}$, the classification accuracy can be defined as:
\begin{equation}
    Accuracy \, = \, \frac{1}{N}\sum_{i=1}^N \mathbbm{1}_{y_i^{pred}=y_i^{true}}
\end{equation}

We also compared the feature representations learned by the models via the t-SNE visualization \cite{van2008visualizing}, and calculated the $\mathcal{A}$-distance \cite{ben2006analysis}, which is commonly used in the DA field to measure the distance between the source and target domains. A smaller $\mathcal{A}$-distance corresponds to a reduced difference between the domains. Finally, following an Explainable Artificial Intelligence (XAI) methodology based on Grad-CAMs \cite{selvaraju2017grad}, we conducted an empirical study to determine the regions within input images that the neural uses to arrive at its final decision. This XAI step helps provide interpretability for medical classification tasks. Algorithm \ref{Ag} presents the pipeline of our evaluation procedure.

\begin{algorithm} \scriptsize
\SetAlgoLined
    \PyComment{Initialize lists to hold performance results} \\
    \PyCode{temp = []} \\
    \PyComment{For each algorithm in the list of algorithms} \\
    \PyCode{for algorithm in Algorithms:} \\
    \Indp
        \PyComment{For each dataset in the list of datasets} \\
        \PyCode{for dataset in Datasets:} \\
        \Indp
            \PyComment{Initialize a temporary list for each dataset} \\
            \PyCode{temp.append([])} \\
            \PyComment{For each epoch up to the specified number of epochs} \\
            \PyCode{for n in range(1, N + 1):} \\
            \Indp
                \PyComment{Compute performance metrics for each epoch} \\
                \PyCode{accuracy, tsne, ... \,=\, compute\_performance\_metrics()} \\
                \PyComment{Append the performance metrics to the temporary list} \\
                \PyCode{temp[-1].append((accuracy, tsne, ...))} \\
            \Indm
            \PyComment{Plot performance curves for each round using the temporary list} \\
            \PyCode{plot\_performance\_curves(temp[-1])} \\
            \PyComment{Use t-SNE to show the source (target) data distributions} \\
            \PyCode{visualize\_data\_distributions\_with\_tsne()} \\
            \PyComment{Calculate the A-distance between source (target) domain} \\
            \PyCode{a\_distance = calculate\_a\_distance()} \\
            \PyComment{Save models \texttt{pth} file using PyTorch utils} \\
            \PyCode{save\_model\_to\_pth()} \\
            \PyComment{Randomly access a sample and use the saved model to visualize the prediction} \\
            \PyCode{visualize\_prediction\_with\_random\_sample()} \\
        \Indm
    \Indm
\caption{General pipeline of our examination}
\label{Ag}
\end{algorithm}


        

\begin{table}[h] \footnotesize
    \setlength{\tabcolsep}{0.87cm}
    \caption{{Description of each dataset.}}
    \begin{tabular}{cccc}
    \toprule
      Datasets&domain&sample(n) &classes  \\
      \midrule
      Office-Home \cite{OfficeHome}&a/c/p/r&2,427/4,365/4,439/4,357&65 \\
      MultiImageNet\cite{hendrycks2021many,wang2019learning}&M/R/S&20,716/24,076/8,152&200 \\
      Office31 \cite{Office31} & A/W/D & 2,817/795/498 & 31 \\
      Adaptiope \cite{Adaptiope}& P/R/S & 12,300/12,300/12,300 &123 \\
      COVID-19 \cite{Gunraj2020} & train/test & 357,518/33,781 & 3\\
      LCOVID-19 \cite{Gunraj2020} & train/test & 300/33,781 & 3\\
 Chest Xray8 \cite{wang2017chestx}& train/test & 86524/25596 & 12\\
      Kidney Cancer \cite{islam2022vision}&train/test & 9000/1000 & 2 \\
  Skin Cancer\footnote{\url{https://www.kaggle.com/datasets/nodoubttome/skin-cancer9-classesisic}} & train/test & 2239/118& 9\\
      Cervical cancer\footnote{\url{https://www.kaggle.com/datasets/obulisainaren/multi-cancer}}&train/test&2500/22500 &5 \\
      Retinal disease \cite{pachade2021retinal}&train/test &1920/640 &2 \\
      Multisource Skin\cite{tschandl2018ham10000,codella2018skin,combalia2019bcn20000}&train/test&22164/6233&7 \\
    \bottomrule
    \end{tabular}
    {a:amazon; c: clipart; p: product; r: real-world; A: Amazon; W: Webcam; D: Dslr; P: Product; P: Real life; S: Synthesis; LCOVID-19: Limited COVID-19.}
    \label{tab:Dataset_Details}
\end{table}

\subsection{Experiments based on natural datasets}

\subsubsection{Mainstream domain adaptation dataset} 

Before we proceed to the simulations, we first present the results of the testing of these algorithms on one of the most popular datasets in the DA field, Office31 \cite{Office31}. Table \ref{tab:Office31} reports the classification accuracy obtained from the DSAN \cite{zhu2020deep} and BNM \cite{BNM} algorithms. The efficacy of these algorithms on the dataset can be clearly seen in the Table. DSAN achieved the highest accuracy (88.4\%), with a large performance gain (+12.3\%). 
The impressive results achieved in this popular benchmarking dataset are the result of various strategies. If we were to test these models on the latest large-scale datasets of computer vision, would they be as effective as for Office31? To answer this question, we initially conducted a test on the Adaptiope dataset \cite{Adaptiope}, which was recently introduced in the field of DA.


\begin{table}[h] \scriptsize
    \setlength{\tabcolsep}{0.57cm}
    \caption{{Summary of the highest classification accuracy (\%) on Office31 and Adaptiope datasets. Except EUDA, all other approaches are based on Resnet50. The best outcomes are highlighted with \textbf{bold} text.}}
    \begin{tabular}{c|ccccccc}
    \toprule
    Office31&A$\rightarrow$W&D$\rightarrow$W& W$\rightarrow$D&A$\rightarrow$D&D$\rightarrow$A&W$\rightarrow$A& Avg \\
    \midrule
Deep Coral&77.7 &97.6 &99.7 &81.1 &64.6 &64.0&80.8  \\
    DANN &82.0 &96.9 &99.1 &79.7 &68.2 &67.4&82.2  \\
    DSAN&\textbf{93.6} &98.3 &\textbf{100} &90.2 &\textbf{73.5} &\textbf{74.8}&\textbf{88.4}  \\
    BNM&91.5 &\textbf{98.5} &\textbf{100} &\textbf{90.3} &70.9 &71.6&87.1  \\
    w/o DA&68.4 &96.7 &99.3 &68.9 &62.5 &60.7& 76.1  \\
    \midrule
    Adaptiope&P$\rightarrow$R&P$\rightarrow$S& R$\rightarrow$P&R$\rightarrow$S&S$\rightarrow$P&S$\rightarrow$R& Avg \\
    \midrule
    Deep Coral&68.4 &36.9 &60.2 &34.7 &46.3 &26.9&39.8  \\
    DANN &69.6 &44.9 &{63.6} &{40.4} &49.9 &27.4&49.3  \\
    DSAN&72.2 &\textbf{67.8} &59.3 &37.6 &51.0 &29.9&53.0  \\
    BNM&{74.9} &61.9 &61.0 &36.8 &{53.7} &{31.7}&{53.3}  \\
    DALN&{79.2}&60.9&{87.6}&{46.9}&{55.9}&{46.9} &{62.9} \\
    DCAN&74.5&45.3&87.3&42.8&54.8&35.1 &56.6 \\
    SACAEM & 79.1&50.7&{89}&\textbf\textbf{48.0}&{58.7}&40.4&61.1 \\
    EUDA &\textbf{87.7}&64.6&\textbf{93.4}&\textbf{60.8}&\textbf{82.4}&\textbf{61.6}&\textbf{75.1} \\
    w/o DA&67.3 &59.3 &62.7 &32.8 &39.9 &23.6&47.6  \\
    \bottomrule
    \end{tabular}
    \label{tab:Office31}
\end{table}

\subsubsection{Simulations on Adaptiope dataset}

Adaptiope dataset is a large-scale dataset containing three domains, namely synthetic (S), product (P) and real-world data (R). Each domain has 123 classes, and the total number of samples in this dataset is 36900 \cite{Adaptiope}. The parameters used for this experiment are listed in Table \ref{T:Details_Alg}. We performed six DA tasks using Resnet50, namely S $\rightarrow$ P, S $\rightarrow$ R, P $\rightarrow$ S, P $\rightarrow$ R, R $\rightarrow$ S, and R $\rightarrow$ P. Table \ref{tab:Office31} reports the classification accuracy using the same DA techniques. {It can be seen that, in terms of overall performance, EUDA achieves the best results (Avg: 75.1\%), followed by DALN (Avg: 62.9\%), SACAEM (Avg: 61.1\%), DCAN (Avg: 56.6\%), BNM (Avg: 53.3\%), DSAN (Avg: 53.0\%), DANN (Avg: 49.3\%), the baseline without DA (Avg: 47.6\%), and Deep Coral (Avg: 39.8\%). These results suggest that for large-scale datasets, correlation techniques such as Deep Coral are less effective for DA. Furthermore, MMD can provide remarkable performance with advanced backbones such as DINOv2, which improves feature representations. We also observe that Deep Coral and DANN suffer from a drop in performance (i.e., from P $\rightarrow$ S), which is commonly known as negative transfer.} Currently, the tests used in our study rely on datasets originating in the DA domain, potentially introducing bias into the obtained results. {Therefore, we chose to use three large-scale challenging dataset, namely ImageNet-M\footnote{\url{https://www.kaggle.com/datasets/ifigotin/imagenetmini-1000/data}}, ImageNet-S \cite{wang2019learning} and ImageNet-R \cite{hendrycks2021many} to further test the algorithms in Subsection \ref{S3.3.3}.}


\subsubsection{Challenging computer vision dataset}\label{S3.3.3}
\begin{figure}
    \centering
    \begin{tabular}{c}
        \includegraphics[width = 0.99 \linewidth]{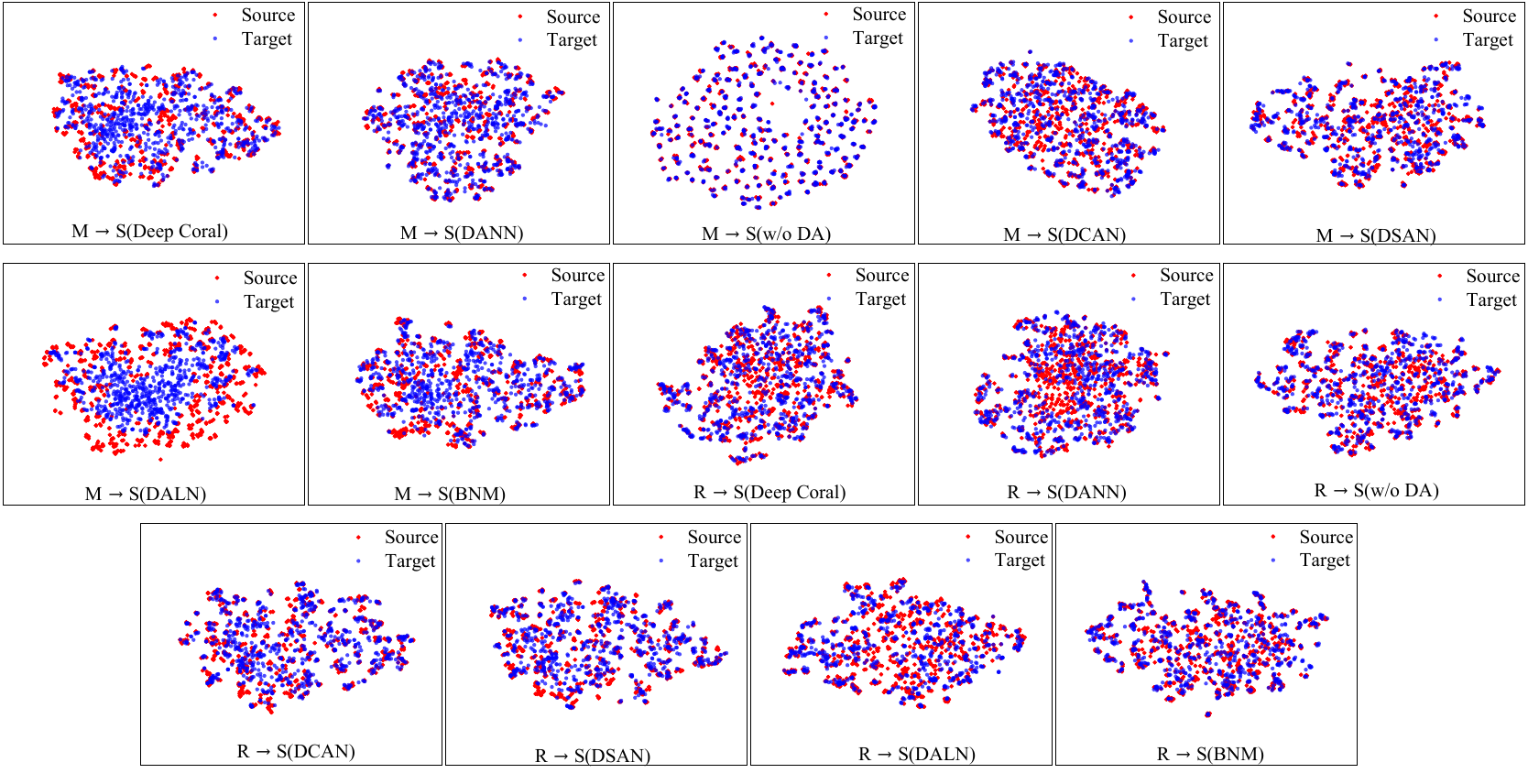} 
    \end{tabular}
    \caption{{The visualization examples of the learned representations using t-SNE for the DA methods using MultiImageNet dataset with Resnet34 as backbone. The red color represents source samples, while the blue color represents target samples.}}
    \label{F:TNSE_ImageNet}
\end{figure}

ImageNet-M is a subset of ImageNet with 1000 classes, each class holds 10\% data in original ImageNet. ImageNet-R is a recent out-of-distribution testing set with 200 classes, which has large feature shifts compared to ImageNet. ImageNet-S is a sketch style dataset, which also varies in texture features compared with ImageNet. Following Office31, six DA tasks were built with Resnet34 as the network backbone. Furthermore, we select the 200 common classes for each domain. The hyperparameters settings are listed in Table \ref{tab:Dataset_Details}. Table \ref{tab:MultiImageNet} reports the best classification accuracy for all methods. {Similar to Adaptiope dataset, Deep Coral shows poor adaptation ability (Avg: 60.4\%), while MMD techniques such as DSAN and DCAN provide higher Avg $\sim 3\%$ compared to Resnet34 alone. We argue that LMMD adaptation captures more dominant invariant features, which yields better performance. In addition, DINOV2 provides rich features, thus using MMD yields 76.6\% Avg. Furthermore, BNM exhibits a lower accuracy in R $\rightarrow$ S and S $\rightarrow$ R than Resnet34 alone. This suggests that adapting only the nuclear norms does not fully leverage the knowledge in the source data. Furthermore, all DA techniques show limited accuracy improvements compared to Resnet34 alone, especially for R $\rightarrow$ \{M, S\} and S $\rightarrow$ \{M, R\}.} {Furthermore, Figure \ref{F:TNSE_ImageNet} shows the t-SNE visualization of the learned feature representations for the source (red) and target (blue) data for the task M $\rightarrow$ S and R $\rightarrow$ S. For M $\rightarrow$ S, the use of DA shows negative adaptation effects, as it shows less clear class boundaries compared to Resnet34 alone. For R $\rightarrow$ S, DCAN and DSAN show better feature alignment compared with DALN and Deep Coral. However, the use of DA shows a limited difference compared to Resnet34 alone. These findings suggest that for large-scale challenging computer vision dataset, the feature adaptation ability of these DA techniques are limited.}

\begin{table}[h] \scriptsize
    \setlength{\tabcolsep}{0.57cm}
    \caption{{Summary of the highest classification accuracy (\%) on MultiImageNet dataset. Except EUDA, all other approaches are based on Resnet34. The best outcomes are highlighted with \textbf{bold} text.}}
    \begin{tabular}{c|ccccccc}
    \toprule
    &M$\rightarrow$R&M$\rightarrow$S& R$\rightarrow$M&R$\rightarrow$S&S$\rightarrow$M&S$\rightarrow$R& Avg \\
    \midrule
    Deep Coral&39.3&49.2&{74.4}&81.2&67.6&51.0 &60.4   \\
    DANN &42.0&53.3&73.2&85.6&67.0&52.5 &62.2   \\
    DSAN&42.3&56.2&72.4&85.1&{68.4}&\textbf{53.5}&63.0   \\
    BNM&40.0&51.7&61.6&61.3&66.6&48.3&54.9   \\
    DALN&34.0&50.7&55.8&\textbf{86.9}&52.8&49.0&54.9  \\
    DCAN&{42.6}&{56.7}&72.8&85.2&68.2&53.3&{63.1}   \\
    EUDA (w/ DINOV2)&\textbf{66.7}&\textbf{72.3}&\textbf{83.4}&83.3&\textbf{84.8}&\textbf{69.1}&\textbf{76.6} \\
    w/o DA&38.4&47.9&72.5&85.8&67.1&51.8&60.6  \\
    \bottomrule
    \end{tabular}
    \label{tab:MultiImageNet}
\end{table}

\begin{figure}[htp]
     \centering
\includegraphics[width = 0.97 \textwidth]{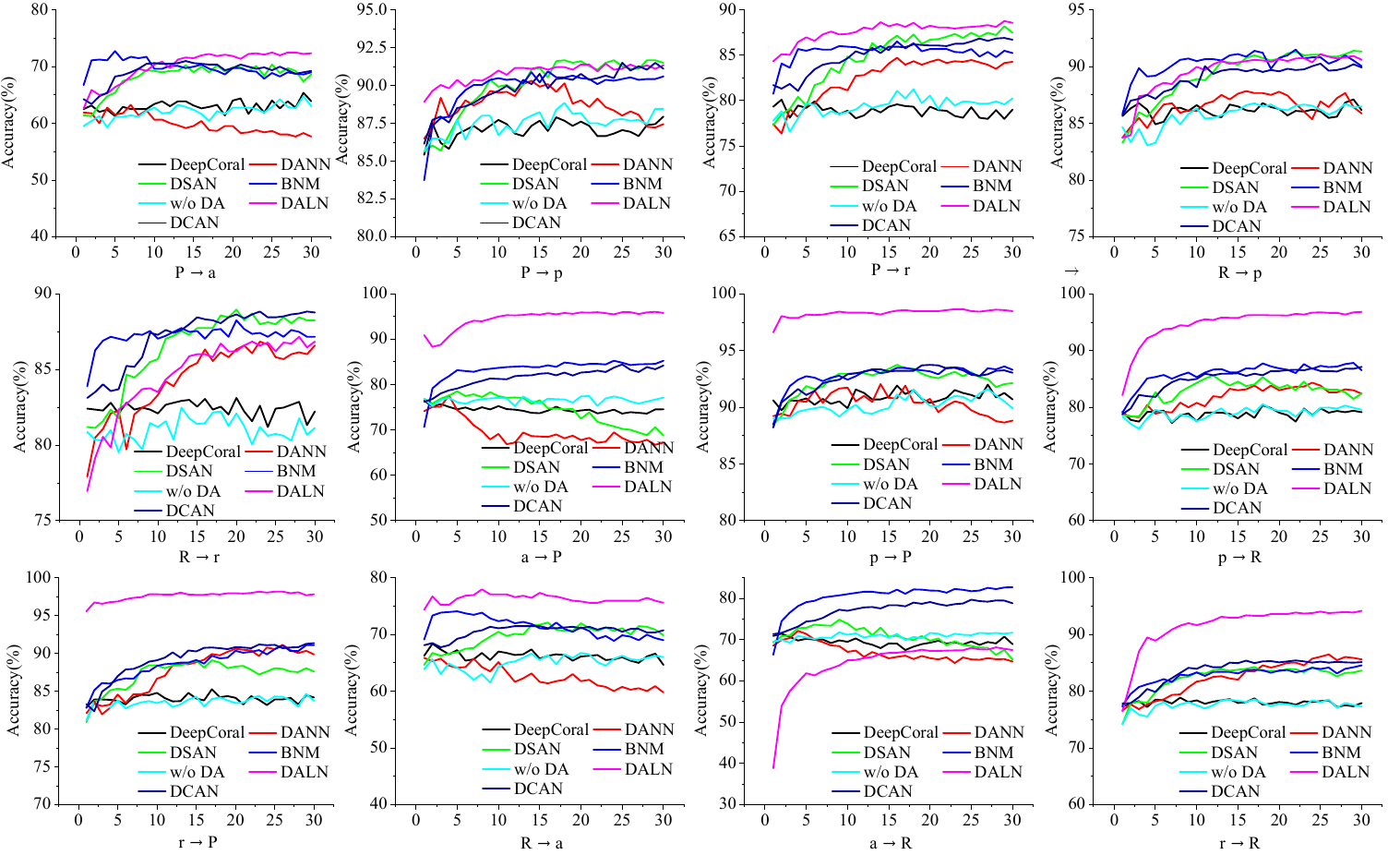}
     \caption{{{Test classification accuracy during training of the DA methods in the cross-dataset with Resnet50 as backbone. The x-axis represents the number of training epochs, while the y-axis shows the classification accuracy.
     }}}
    \label{F:Cross-Dataset}
\end{figure}

\subsubsection{Cross-dataset environment} 
Next, we evaluated the cross-dataset robustness of algorithms by training a model on a given dataset and then testing it on another dataset, using common labels from both datasets. For example, we trained our model using the Office31 dataset and evaluated its performance by testing it on the Office-Home dataset. Conducting such experiments is important as most advancements in the field, based on mainstream datasets such as Office31 and Office-Home, use domains within the same dataset.  We considered 12 domain adaptation tasks using the Adaptiope (P, R) and Office-Home (a, p, r) datasets. These tasks include P $\rightarrow$ a, P $\rightarrow$ p, P $\rightarrow$ r, R $\rightarrow$ a, R $\rightarrow$ P, R $\rightarrow$ r, a $\rightarrow$ P, a $\rightarrow$ R, p $\rightarrow$ P, p $\rightarrow$ R, r $\rightarrow$ P, and r $\rightarrow$ R. Table \ref{tab:cross-dataset_} reports the classification results of the DA algorithms for these tasks. The classification accuracy of the algorithms, measured at each training epoch, is shown in Figure \ref{F:Cross-Dataset}. {As illustrated, Deep Coral leads to minimal accuracy enhancement, consistent with the findings on the Adaptiope dataset. Furthermore, DANN gives a slight increase in accuracy 2\%. {Moreover, DSAN accuracy notably increases by 4.4\%, while BNM average accuracy improves by an additional 6.1\%.} Furthermore, EUDA has the highest average test accuracy (92.4\%). Thus, adversarial based techniques such as DALN and advanced backbone based approaches such as EUDA are robust to cross-dataset situation as it can capture rich domain-invariant features.}

\begin{table*} \scriptsize
    \caption{{Highest classification accuracy (\%) obtained by DA methods in the cross-dataset. Except EUDA, all other approaches are based on Resnet50. The best outcomes are highlighted with {bold} text.}}
    \setlength{\tabcolsep}{0.2cm}
    \begin{tabular}{c|ccccccccccccc}
    \toprule
     Alg. &P$\rightarrow$a &P$\rightarrow$p &P$\rightarrow$r &R$\rightarrow$a &R$\rightarrow$p &R$\rightarrow$r &a$\rightarrow$P &a$\rightarrow$R &p$\rightarrow$P &p$\rightarrow$R &r$\rightarrow$P &r$\rightarrow$R&Avg  \\
     \midrule
     DC   &65.4 &87.9 &80.1 &68.4 &87.1 &83.1 &76.5 &71.4 &92.0 &80.3 &85.2 &78.8& 79.7 \\
     DANN &63.3 &90.3 &84.7 &65.9 &87.8 &86.8 &77.6 &72.1 &92.0 &84.3 &{91.1} &{86.5}& 81.9 \\
     DSAN &70.4 &{91.7} &{88.2} &72.1 &91.3 &{{88.9}} &78.5 &74.8 &{93.7} &85.6 &89.1 &84.2& 84.0 \\
     BNM &{72.7} &90.9 &86.5 &{74.1} &{91.5} &88.2 &{85.3} &{82.7} &93.6 &{87.8} &{91.1} &84.5& {85.7} \\
     DALN&72.5&91.4&{88.8}&{78.0}&91.1&87.2&\textbf{95.8}&68.1&\textbf{98.7}&{96.8}&\textbf{98.2}&\textbf{94.2}&{88.4} \\
     DCAN&71.0&{91.5}&86.9&71.5&90.3&88.8&84.4&79.7&93.7&87.1&91.3&85.4&85.1  \\
     EUDA (w/ DINOV2)&\textbf{84.8}&\textbf{95.7}&\textbf{92.1}&\textbf{82.9}&\textbf{95.9}&\textbf{92.4}&91.3&\textbf{91.3}&97.3&\textbf{97.1}&91.9&94.1&\textbf{92.2} \\
     w/o DA &64.8 &88.8 &81.2 &66.7 &86.8 &82.5 &77.5 &71.8 &91.6 &80.5 &84.6 &78.5& 79.6 \\
     \bottomrule
    \end{tabular}
    \label{tab:cross-dataset_}
\end{table*}

\subsection{Experiments based on medical datasets} 

{\subsubsection{Datasets} To assess the usefulness of DA methods in medical imaging applications, we chose seven datasets based on their size, which ranged from a few thousand to tens of thousands of samples. The following sections provide a detailed description of each dataset. Further information can be found in Table \ref{tab:Dataset_Details}.}

{\mypar{ChestXray8} The ChestXray8 dataset was chosen for analysis \cite{wang2017chestx}. This dataset contains 108948 X-ray images from 32717 unique patients with 12 disease image labels (Atelectasis (1), Cardiomegaly (2), Effusion (3), Infiltration (4), Mass (5), Nodule (6), Pneumonia (7), Pneumothorax (8), Consolidation (9), Edema (10), Emphysema (11), and Fibrosis (12)).}

{\mypar{COVID-19}
COVID-19 dataset \cite{Gunraj2020} is a large-scale medical dataset, which has a training set comprised of 357518 samples and a testing set consisting of 33781 images. It has three classes, normal (0), pneumonia (1), and COVID-19 (2).}

{\mypar{Kidney cancer} The dataset consists of 10000 CT-radiography images of kidneys with two distinct labels, tumor and normal \cite{islam2022vision}, with an equal distribution of 5000 samples per class. As there is no internal testing set, we randomly selected samples from the data for evaluation. Specifically, we allocated 90\% of the samples for training purposes and the remaining 10\% for testing.}

{\mypar{Skin cancer} The Skin Cancer ISIC dataset which comprises 2357 images of malignant and benign skin lesions\footnote{\url{https://www.kaggle.com/datasets/nodoubttome/skin-cancer9-classesisic}}. There are nine classes in this dataset, namely actinic keratosis, basal cell carcinoma, dermatofibroma, melanoma, nevus, pigmented benign keratosis, seborrheic keratosis, squamous cell carcinoma and vascular lesion.}

{\mypar{Cervical cancer} This dataset consists of 25000 images related to cervical cancer\footnote{\url{https://www.kaggle.com/datasets/obulisainaren/multi-cancer}}, which are divided into five classes: Dyskeratotic, Koilocytotic, Metaplastic, Parabasal, and Superficial-Intermediate. In this particular dataset, our objective was to evaluate the capacity to generalize the knowledge acquired from a limited sample to a larger domain. Specifically, we randomly selected 10\% samples independently of each class for training purposes, and used the rest of samples for testing.}

{\mypar{Retinal disease} This dataset comprises a total of 1920 images for training and 640 images for testing \cite{pachade2021retinal}, and consists of two distinct classes, namely healthy and unhealthy.}

{\mypar{Multisource skin cancer} This dataset consists of three distinct skin cancer dataset collected from three different sources (kaggle\footnote{\url{https://www.kaggle.com/datasets/nodoubttome/skin-cancer9-classesisic/data}}, HAM10000 \cite{tschandl2018ham10000} and ISIC2019 \cite{combalia2019bcn20000,codella2018skin}). We select the common classes in these datasets (i.e., actinic keratosis (AK), basal cell carcinoma (BCC), dermatofibroma (DF), melanoma (MEL), nevus (NV), pigmented benign keratosis (PBK), and vascular lesion (VL)). Since ISIC2019 lacks a test set, we divided it into two parts, one for training and the other for testing, with a ratio of 8:2. {Furthermore, the test sets provided in kaggle and HAM10000 datasets are combined together with the ISIC2019 test set for testing.}


{As before, we used the Resnet34 \cite{Resnet}, Resnet50 \cite{Resnet} and Densenet121 \cite{huang2017densely} architectures as backbone network. The same networks were considered for the kidney cancer, skin cancer, cervical cancer, retinal disease and multisource skin cancer. For ChestXray8, Resnet34 was replaced with Resnet101. Due to the class imbalance in multisource skin cancer, we considered balanced accuracy (mean value of each class recall) and macro f1 score for evaluation.}

\begin{figure}[!ht]
    \centering
     \includegraphics[width = 0.97 \textwidth]{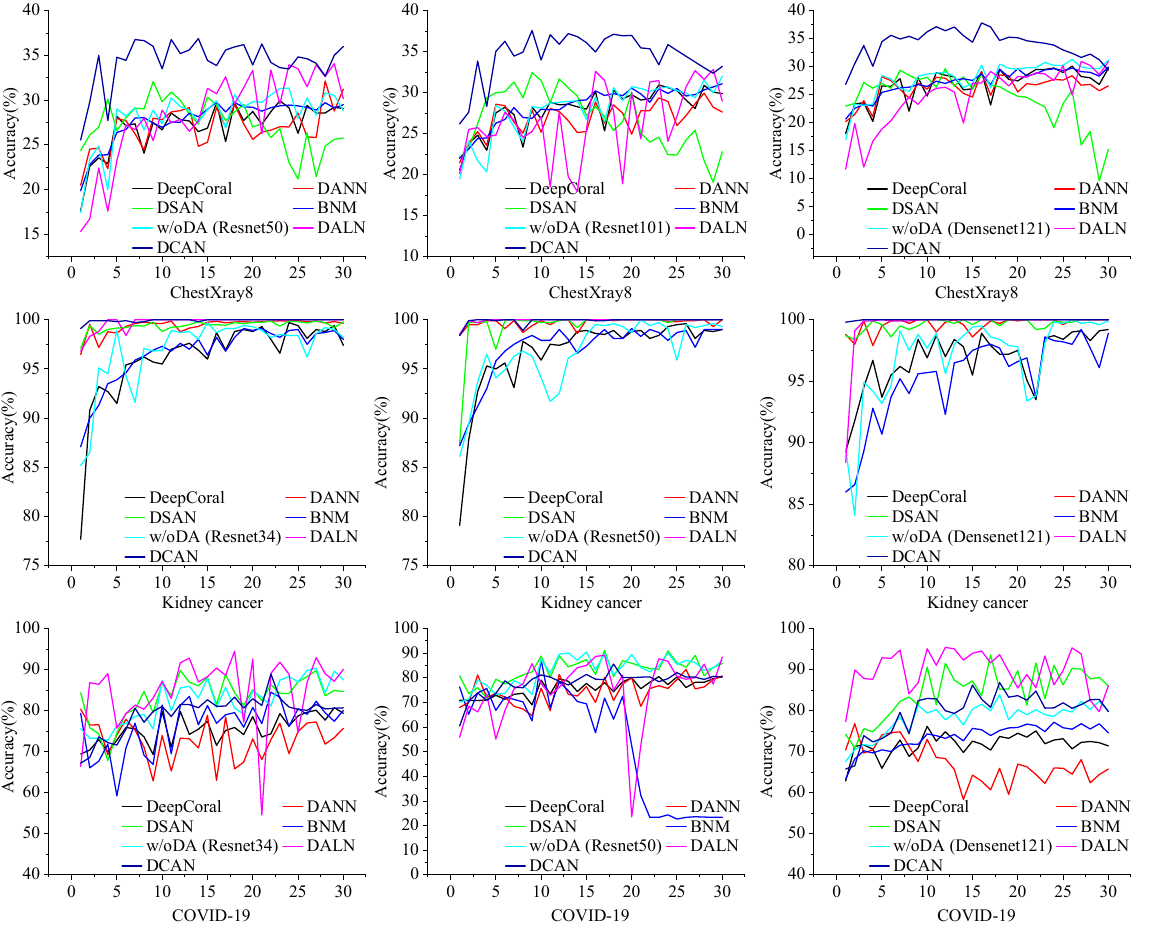}
   
    \caption{{Test classification accuracy obtained using six DA techniques on medical datasets. The x-axis represents the number of training epochs. The columns, from left to right, represent Resnet34, Resnet50, and Densenet121, respectively.
    }}
    \label{Results_All_medical}
\end{figure}

\subsubsection{Results}  

\mypar{ChestXray8}  For this dataset, all training parameters were the same as in Table \ref{T:Details_Alg} except for the number of iterations per epoch that we set to 1000. Figure \ref{Results_All_medical} shows the classification accuracy during the training of the DA algorithms. {For example, BNM, Deep Coral and DANN exhibit poor performance in the three backbone networks compared to not using any DA. However, the use of DALN and DCAN leads to an increase in accuracy with Resnet50 and Resnet101, especially for DCAN with an increase in accuracy $> 5\%$.} Furthermore, in this case, DSAN shows severe overfitting with its accuracy starting to decline around epoch 17 and dropping to 10-15\% at the end of training. A similar trend of overfitting is also observed for Resnet101. 
The performance results of tested methods are summarized in Table \ref{tab:ChestXray}. Again, we observe that methods have limited efficacy in addressing this challenging task. {We argue that the abnormal regions in chest x-ray images only represent a small part of the image, while the rest are nearly identical, it is challenging to extract useful features related to tasks, thus limiting the potential of these DA techniques.}

\mypar{Kidney cancer}
The parameters used for this dataset are the same as in the Table \ref{T:Details_Alg}. The classification accuracy of the algorithms, measured at each training epoch, is shown in Figure \ref{Results_All_medical}. {DANN, DSAN, DALN and DCAN obtain an outstanding testing accuracy from early stages of training, whereas Deep Coral and BNM require a higher number of epochs to achieve comparable performance.} In general, both DANN and DSAN yield better results compared to the same CNN without DA. The final performance metrics of these algorithms are reported in Table \ref{tab:summary}. {Since without DA yields remarkable performance (e.g., 99.2\% accuracy), the role of DA here mainly focuses on improving early stage performance (i.e., with DA requires less time to achieve comparable accuracy).}

\mypar{COVID-19} For the COVID-19 dataset, we used the same parameters as those in Table \ref{T:Details_Alg}, except for the iterations per epoch parameter which was set to 300. Figure \ref{Results_All_medical} plots the evolution of accuracy during training of these techniques. As can be observed, the BNM algorithm exhibits important overfitting for the Resnet50 backbone, with its accuracy dropping below 30\% during training. Furthermore, in many cases, DA algorithms did not achieve any improvement over the CNN baseline without adaptation (w/o DA). Except for DSAN, DALN, and DCAN on Densenet121, the DA algorithms did not show positive results. Table \ref{tab:summary} reports the classification accuracy (highest) using DA and CNN alone on the COVID-19 dataset. {For example, DALN achieves the highest testing accuracy using Resnet34 and Densenet121, while DSAN shows the best testing accuracy of 91.2\%.} However, Deep Coral and DANN illustrate inferior accuracy compared to CNN alone. {These results suggest that correlation based techniques are not useful to adapt the CT scan images, while MMD and adversarial based approaches are more suitable.}

\begin{table}[htp] \scriptsize
    \setlength{\tabcolsep}{1.32cm} 
    \caption{{Top classification accuracy (\%) on the ChestXray8 dataset. The highest results are indicated in \textbf{bold} text.} }
    \begin{tabular}{c|ccc}
    \toprule
    Alg.& \multicolumn{3}{c}{Accuracy} \\
    \midrule
    & Resnet50 & Resnet101 & Densenet121 \\
    \midrule
    Deep Coral&29.9&30.9&30.1 \\
    DANN &31.2&29.9&29.0 \\
    DSAN&{32.0}&31.7&29.6  \\
    BNM&29.7&31.1&29.9  \\
    DALN&34.1 &32.8 &30.9  \\
    DCAN &\textbf{36.9} &\textbf{37.6} &\textbf{37.8}  \\
    w/o DA&31.3&{31.9}&{31.3}   \\
    \midrule
    EUDA (w/ DINOV2) &\multicolumn{3}{c}{35.6}\\
    \bottomrule
    \end{tabular}
    \label{tab:ChestXray}
\end{table}

\begin{table}[htp] \scriptsize
    \setlength{\tabcolsep}{0.97cm}
    \caption{{Summary of the highest classification accuracy (\%) on medical datasets. The best results are highlighted with \textbf{bold} text.}}
    \begin{tabular}{cc|ccc}
    \toprule
    &Alg.& \multicolumn{3}{c}{Accuracy} \\
    \midrule
    & &Resnet34 & Resnet50 & Densenet121 \\
    \midrule
  \multirow{7}{*}{\rotatebox{90}{COVID-19}}&Deep Coral&80.7&80.5&75.1 \\
           & DANN &80.4&83.4&76.8 \\
    &DSAN&89.8&\textbf{91.2}&{91.7}  \\
    &BNM&84.5&87.5&77.2  \\
    &DALN&\textbf{94.5} &89.2 &\textbf{95.4} \\ 
    &DCAN &88.9 &85.6 &86.9  \\
    &w/o DA&{90.4}&90.5&83.9   \\
    \midrule
    \multirow{7}{*}{\rotatebox{90}{Kidney cancer}}&Deep Coral&99.7&99.6&99.2 \\
    &DANN &{99.9}&\textbf{100}&\textbf{100} \\
    &DSAN&{99.9}&\textbf{100}&\textbf{100}  \\
    &BNM&98.9&99.0&99.2  \\
    &DALN&\textbf{100} &\textbf{100} &\textbf{100} \\
    &DCAN &\textbf{100} &\textbf{100} &\textbf{100}  \\
    &w/o DA&99.2&\textbf{100}&\textbf{100} \\
    \midrule
    \multirow{7}{*}{\rotatebox{90}{Skin cancer}}&Deep Coral&59.3&61.0&61.9 \\
    &DANN &55.9&58.5&62.7 \\
    &DSAN&\textbf{61.9}&\textbf{67.8}&61.9  \\
    &BNM&59.3&60.2&\textbf{65.3}  \\
    &DALN&58.6 &64.6 &59.6 \\
    &DCAN &55.9 &61.0 &61.8  \\
    &w/o DA&59.3&61.0&62.7 \\
    \midrule
    \multirow{7}{*}{\rotatebox{90}{Cervical cancer}}&Deep Coral&98.1&98.2&97.3 \\
    &DANN &98.5&{98.7}&{98.2} \\
    &DSAN&{98.6}&98.6&98.1  \\
    &BNM&96.2&96.9&96.6  \\
    &DALN&99.4 &99.5 &99.7 \\
    &DCAN &\textbf{99.8} &\textbf{99.8} &\textbf{99.9}  \\
    &w/o DA&97.8&98.4&97.7 \\
    \midrule
    \multirow{7}{*}{\rotatebox{90}{Retinal disease}}&Deep Coral&91.3&92.0&{91.4} \\
    &DANN &91.3&{92.8}&90.8 \\
    &DSAN&{92.0}&91.4&90.3  \\
    &BNM&91.7&90.9&90.5  \\
    &DALN&91.9 &\textbf{93.3} &\textbf{93.0} \\
    &DCAN &\textbf{92.3} &91.9 &91.4  \\
    &w/o DA&91.6&91.4&90.3 \\
    \midrule
    \end{tabular}
    \setlength{\tabcolsep}{0.4cm}
    \begin{tabular}{c|ccccc}
      Alg.   & COVID-19 & Kidney cancer & Skin cancer & Cervical cancer & Retinal disease  \\
    \midrule
       EUDA (with DINOv2) & 88.5&99.4 &52.5 &98.5 &92.1  \\
    \bottomrule
    \end{tabular}
    \label{tab:summary}
\end{table}

\mypar{Skin cancer} Figure \ref{F:Results_All2} shows the classification accuracy of the algorithms tested during the training procedure. DSAN gives the highest performance, reaching 61.9\%, when employing Resnet34.
Additionally, BNM shows slightly superior performance compared to DANN and Deep Coral, but still falls short of DSAN. Only DSAN produces superior results compared to the baseline without DA. Furthermore, DSAN also gives the highest accuracy when using Resnet50 as a backbone, whereas BNM leads to the highest improvement in the case of Densenet121. {The DSAN, DCAN, DALN, and Deep Coral methods are less effective than the CNN-only baseline when applied to Densenet121. However, BNM demonstrates enhanced performance.} The performance of each method in this data set is summarized in Table \ref{tab:summary}. Figure \ref{F:TNSE_Medical_All} gives a t-SNE plot of the representations learned by the DA methods for the skin cancer dataset. This analysis reveals that BNM is less effective in addressing discrepancies in the data distribution, compared to the other three algorithms. DSAN and DCAN algorithms demonstrate superior performance for target data alignment, showing distinct boundaries between data clusters. Moreover, Deep Coral exhibits better results compared to BNM, and a performance on par with that of DANN. This observation further demonstrates the effectiveness of MMD-based methodologies in reducing the differences between the source and target domains.  {Overall, although MMD based approaches can solve the class imbalance in skin cancer dataset (more reliable class clusters), these techniques may adapt the features of rare class samples to the wrong clusters (negative transfer), thereby reduces the test performance compared to adversarial DA methods.}



\begin{figure}
    \centering
     \includegraphics[width = 0.99 \linewidth]{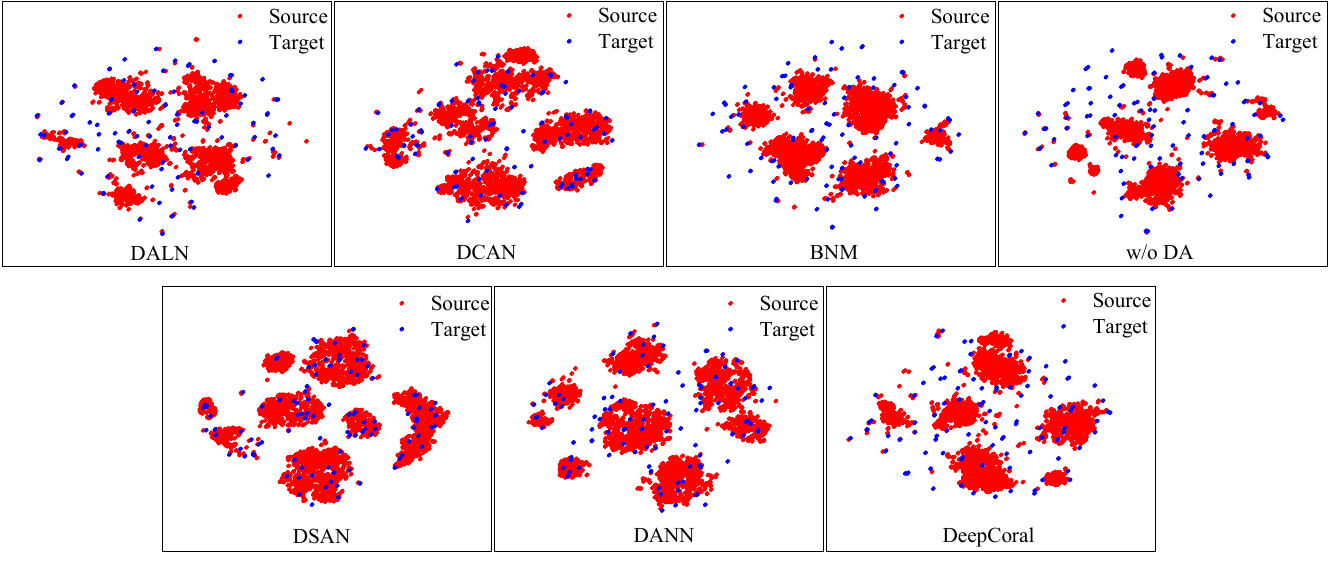}
    \caption{{The visualization examples of the learned representations using t-SNE for the DA methods using Skin cancer dataset with Resnet50 as backbone. The red color represents source samples, while the blue color represents target samples.}}
    \label{F:TNSE_Medical_All}
\end{figure}

\mypar{Cervical cancer} We use a cervical cancer dataset for our analysis. Table \ref{tab:summary} reports the highest classification accuracy obtained by the DA algorithms, and Figure \ref{F:Results_All2} shows the classification accuracy at each training epoch. These results indicate that the six DA algorithms exhibited similar overall performance during the training phase, compared to the baseline without DA. Moreover, all algorithms experienced slight improvements when Resnet34 was used as a backbone, except BNM, which obtained a lower accuracy than the CNN-only baseline. However, when employing Resnet50 and Densenet121 as a backbone, both Deep Coral and BNM produce a lower performance than this baseline. {Furthermore, DCAN demonstrates the most stable performance ($>99.5\%$ accuracy) using many network backbones.} {Overall, similar to kidney cancer, MMD and adversarial based techniques are more effective compared to correlation based approach, and the main ability of DA is to reduce the training epochs to reach its best accuracy.}




\mypar{Retinal disease} {The parameters used for this dataset correspond to those presented in Table \ref{T:Details_Alg}.} Table \ref{tab:summary} reports the maximum classification accuracy achieved by DA techniques, and Figure \ref{F:Results_All2} plots the accuracy measured throughout training. We find that the accuracy of the six DA algorithms fluctuates in a range of 80\% to 95\% during training. Moreover, when using Resnet34 as backbone, DA algorithms only obtain marginal improvements. Deep Coral and DANN even exhibit lower accuracy than the baseline without DA. In contrast, when employing Resnet50, Deep Coral and DANN achieved a superior performance compared to the CNN-only baseline; however, the remaining methods showed a slightly reduced performance. Lastly, when Densenet121 is used, most algorithms obtained a partial improvement, except DSAN, which achieved a score similar to that without DA. Among the tested algorithms, DALN showed the highest improvement, with an increase in accuracy of 2.7\%. {These results suggest that adversarial DA methods are robust to retinal diseases, while correlation and MMD based techniques can not fully adapt the task-related features.}

\begin{table}[!ht] \scriptsize
    \centering
    \setlength{\tabcolsep}{13.5pt}
        \caption{{Summary of the classification accuracy (\%), balanced accuracy (\%) and macro-F1 (\%) using multisource skin cancer dataset. The best outcomes are highlighted with \textbf{bold} text.}}
    \begin{tabular}{c|ccc|ccc|ccc}
    \toprule
    &\multicolumn{3}{c|}{Resnet34}&\multicolumn{3}{c|}{Resnet50}&\multicolumn{3}{c}{Densenet121} \\
    Alg.&ACC&BACC&F1&ACC&BACC&F1&ACC&BACC&F1  \\
    \midrule
    Deep Coral&87.2 &80.2 &81.5 &88.2 &82.2 &82.7 &88.2 &80.7 &82.5 \\
    DANN &88.8&\textbf{84.3}&84.1&89.5&84.1&84.7&\textbf{89.6}&83.3&84.8 \\
    DSAN &88.7&83.7&83.8&90.4&85.9&\textbf{86.8}&89.3&\textbf{84.8}&85.1 \\
    BNM &80.1&74.3&67.4&82.3&78.4&69.1&82.3&77.2&70.3 \\
    DALN &87.4&83.2&80.7&\textbf{90.6}&\textbf{87.5}&85.4&86.7&80.5&81.0 \\
    DCAN &\textbf{89.0}&83.9&\textbf{84.6}&90.4&85.8&86.1&89.4&\textbf{84.8}&\textbf{85.5} \\
    w/o DA &88.7&83.7&83.8&90.4&85.9&\textbf{86.8}&89.3&\textbf{84.8}&85.1 \\
    
    \midrule
    \end{tabular}
\setlength{\tabcolsep}{41pt}
\begin{tabular}{c|ccc|ccc|ccc}
   Alg. & \multicolumn{3}{c}{ACC}& \multicolumn{3}{c}{BACC}&\multicolumn{3}{c}{F1}  \\
\midrule
   EUDA (w/ DINOV2) & \multicolumn{3}{c}{81.9} & \multicolumn{3}{c}{69.7} & \multicolumn{3}{c}{72.7} \\ 
\bottomrule
\end{tabular}

    \label{tab:RealSkin}
\end{table}

{\mypar{Multisource skin cancer} Table \ref{tab:RealSkin} reports the highest testing ACC, BACC and F1 for all methods. The use of DA shows limited potential to improve ACC, BACC, and F1. BNM shows lower ACC, BACC, and F1 compared to CNN alone. {Furthermore, the use of BNM leads to ACC, BACC, and F1 score decrease (e.g., $>5\%$ drop compared with Densenet121 alone).} This suggests that DA is not very useful for large-scale imbalanced skin cancer classification datasets.} {Furthermore, Figure \ref{F:TNSE_RealSkin} reports the t-SNE visualization of learned features from source and target data using Resnet50. The use of DCAN and DSAN leads to clearer class boundaries compared to Deep Coral, DALN, and Resnet50 alone, demonstrating the potential of MMD-based feature alignment for large-scale skin cancer data.} {In summary, MMD-based techniques are more suitable compared to correlation and adversarial-based approaches. However, due to the class imbalance in skin cancer images, the effectiveness of all DA methods is limited. This indicates that feature shifts in skin cancer are not the primary challenge affecting model performance.}

\begin{figure}
    \centering
     \includegraphics[width = 0.99 \linewidth]{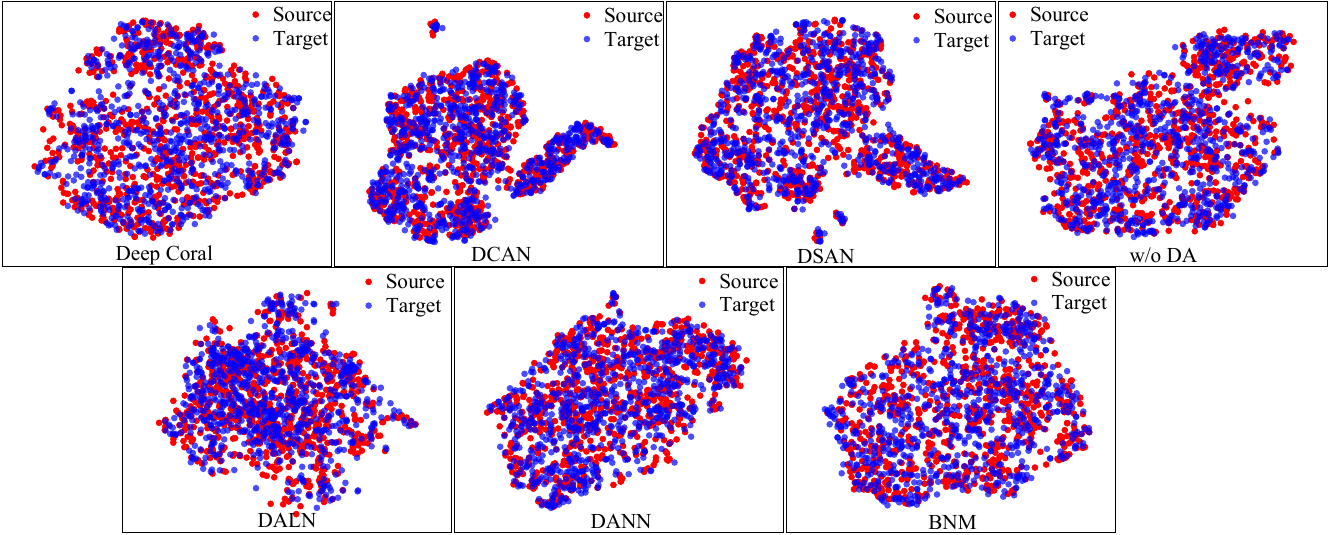}
    \caption{{The visualization examples of the learned representations using t-SNE for the DA methods using multisource skin cancer dataset with Resnet50 as backbone. The red color represents source samples, while the blue color represents target samples.}}
    \label{F:TNSE_RealSkin}
\end{figure}

\begin{figure}
    \centering
     \includegraphics[width = 0.96 \textwidth]{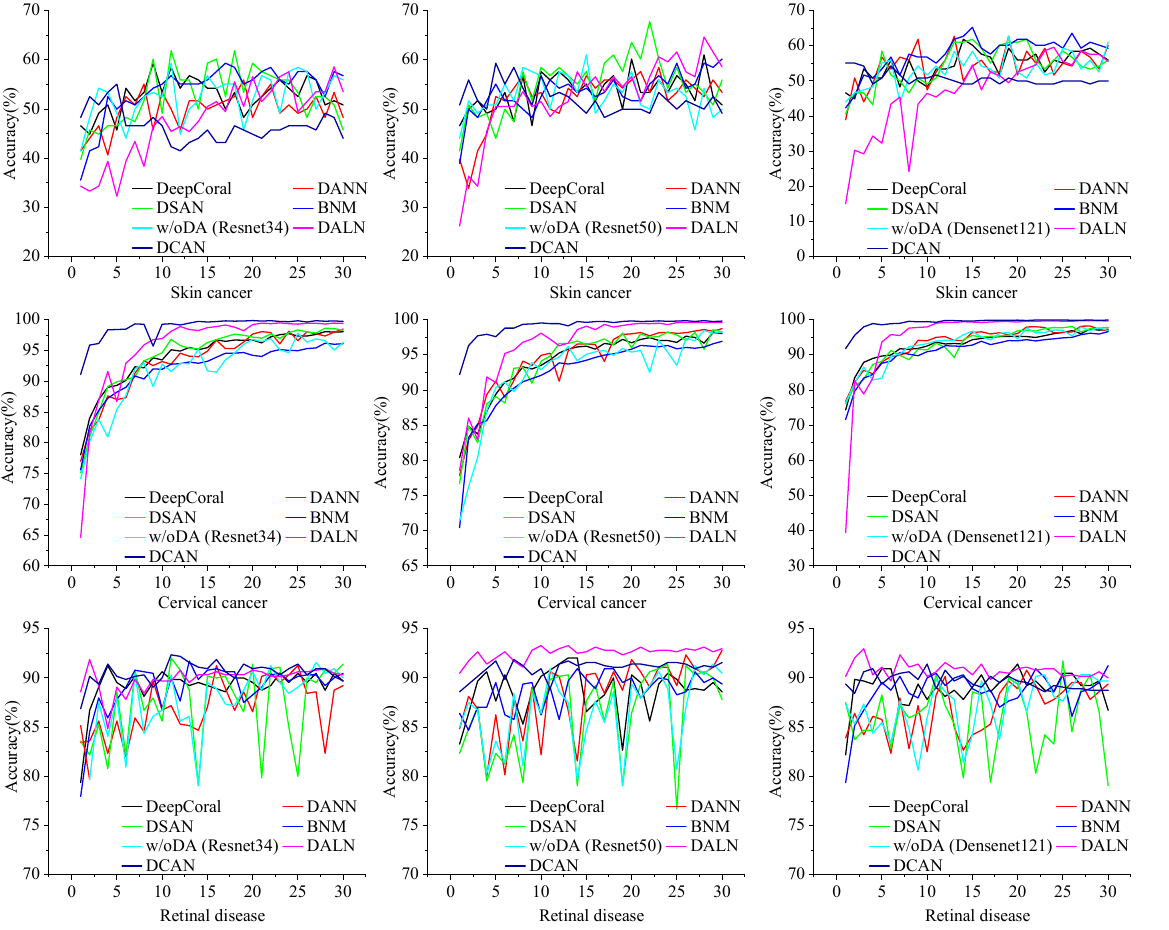}
    \caption{{Test classification accuracy obtained using six DA techniques on medical datasets. The x-axis represents the number of training epochs. The columns, from left to right, represent Resnet34, Resnet50, and Densenet121, respectively.}}
    \label{F:Results_All2}
\end{figure}

\section{{Exploring the domain adaptation techniques under various situations}}\label{S:4}

\subsection{The impact of different batch size}
Processing larger batches of data often presents a significant challenge for embedded devices and edge devices due to the need for powerful GPUs. In this regard, we are interested in knowing how the DA algorithms can perform with smaller batches. Thus, we examine the performance of these algorithms using three different batch sizes: 4, 8, and 16 (baseline). For this experiment, a natural dataset (Adaptiope) and a medical dataset (COVID-19) were selected for testing. The parameters used for this experiment correspond to those presented in Table \ref{T:Details_Alg}, except for the batch size which varies as mentioned above. We chose Resnet50 as our backbone for testing. Table \ref{tab:adaptiope} reports the performance of the algorithms tested for the Adaptiope dataset. {DALN demonstrates higher accuracy compared to Resnet50 alone. Furthermore, the use of DSAN shows almost identical accuracy compared to Resnet50 only with a batch size of 4.} {These results suggest that LMMD is less effective compared to correlation and adversarial based techniques. We contend that the constraint of limited batch size results in fewer samples from diverse classes and reduces the potential of LMMD.}

\textcolor{black}{The best accuracy of methods for the COVID-19 dataset is reported in Table \ref{tab:COVID-BS}. As observed, Deep Coral obtains substantial improvements when smaller batches are used. However, severe performance degradation is found for BNM, and DSAN also experiences a decline in accuracy. For these two algorithms, using a smaller batch size may result in a noisy DA procedure which is specific to each batch. This increased variability across batches can lead to instability that eventually degrades performance. In comparison, we find that DANN and the baseline without DA are more robust in terms of batch size. }

\begin{table}[h] \scriptsize
    \setlength{\tabcolsep}{0.6cm}
    \caption{{Summary of the highest classification accuracy (\%) on Adaptiope datasets using two batch sizes with Resnet50 as backbone. The best outcomes are highlighted with \textbf{bold} text.}}
    \begin{tabular}{c|ccccccc}
    \toprule
Batch size\,=\,8&P$\rightarrow$R&P$\rightarrow$S& R$\rightarrow$P&R$\rightarrow$S&S$\rightarrow$P&S$\rightarrow$R& Avg \\
    \midrule
    Deep Coral& 64.8& {33.8}& {85.9}& {34.9}& {46.2}& 26.0& {48.6}  \\
    DANN & 60.6& 28.6& 85.7& 30.6& 38.4& 20.6& 44.1  \\
    DSAN& 59.8& 30.9& 84.1& 30.4& 41.7& 18.8& 44.3  \\
    BNM& {71.2}& 33.0& {85.9}& 28.3& 40.1& {26.6}& 47.5  \\
    DALN&\textbf{74.3}&\textbf{48.1}&\textbf{88.6}&\textbf{47.8}&\textbf{55.1}&\textbf{35.7}&\textbf{58.2} \\
    DCAN&66.4&38.1&84.3&35.5&42.8&22.1&48.2  \\
    w/o DA& 54.3& 27.1& 81.7& 25.9& 34.3& 18.8& 40.3  \\
\midrule
Batch size\,=\,4&P$\rightarrow$R&P$\rightarrow$S& R$\rightarrow$P&R$\rightarrow$S&S$\rightarrow$P&S$\rightarrow$R& Avg \\
    \midrule
    Deep Coral&59.6 &{32.5} &75.9 &{32.1} &\textbf{43.8} &{25.4}&{44.9}  \\
    DANN &50.3 &22.8 &72.2 &22.6 &31.1 &13.6&35.4  \\
    DSAN&40.1 &15.2 &61.9 &17.6 &21.4 &12.0&28.0 \\
    BNM&{63.7} &28.5 &{76.7} &20.7 &36.9 &24.0&41.8  \\
    DALN&64.2&33.1&\textbf{83.0}&31.5&35.9&16.0&43.9 \\
    DCAN&\textbf{67.6}&\textbf{36.6}&82.1&\textbf{33.4}&41.9&\textbf{28.7}&\textbf{48.3}  \\
    w/o DA&40.6 &13.7 &61.9 &17.2 &21.3 &12.5&27.9  \\
    \bottomrule
    \end{tabular}
    \label{tab:adaptiope}
\end{table}



\begin{table}[h] \scriptsize
    \setlength{\tabcolsep}{1.2cm}
    \caption{{Summary of the highest classification accuracy (\%) on COVID-19 datasets with Resnet50 as backbone. The best outcomes are highlighted with \textbf{bold} text.}}
    \begin{tabular}{c|ccc}
    \toprule
    Alg.& \multicolumn{3}{c}{Accuracy}\\
    \midrule
    & Batch size\,=\,16 & Batch size\,=\,8 & Batch size\,=\,4 \\
    \midrule
    DC &80.5 &81.8 &90.5 \\
    DANN &83.4 &84.2 &84.8 \\
    DSAN &\textbf{91.2} &88.7 &83.3 \\
    BNM &78.4 &69.7 &42.5 \\
    DALN&89.2 &\textbf{94.5} &85.1   \\
    DCAN&85.6 &87.1 &77.6  \\
    w/o DA &90.5 &{90.8} &\textbf{91.3}  \\
    \bottomrule
    \end{tabular}
    \label{tab:COVID-BS}
\end{table}

\begin{figure}[htp]
     \centering
    \includegraphics[width = 0.98 \textwidth]{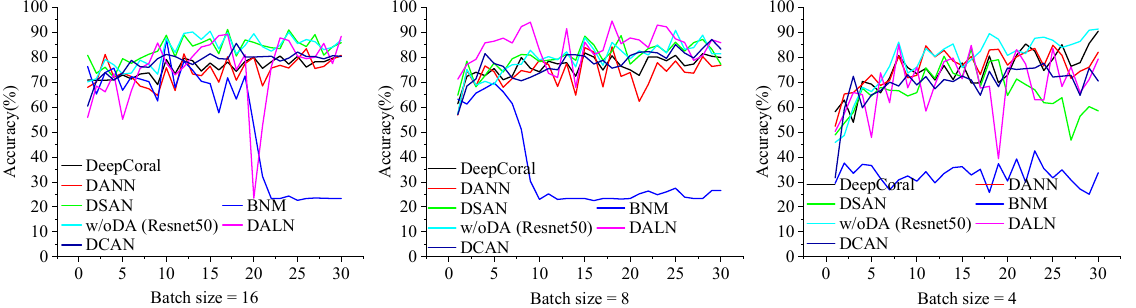}
    \caption{{Test classification accuracy using six DA techniques on COVID-19 dataset with three batch sizes. The columns, from left to right, represent Resnet34, Resnet50, and Densenet121, respectively. The x-axis represents the number of training epochs.}}
    \label{F:COVID_BatchSize}
\end{figure}

\textcolor{black}{\subsection{Limited GPU memory} }

{In previous experiments, we investigated the effectiveness of DA algorithms when employing standard networks such as Resnet50. In the next experiment, we assess the ability of these algorithms to generalize under computational power constraints (e.g., low-end GPU with only 1 or 2 GB of memory) by selecting two memory/computation-efficient neural networks: ShuffleNet \cite{zhang2018shufflenet} and MobileNet \cite{sandler2018mobilenetv2}. We analyze the performance of algorithms using these efficient backbones for three different batch sizes: 4, 8, and 16. Table \ref{tab:COVID_Limit_Computation} reports the results of this experiment. Notable performance discrepancies are observed with a batch size of 4. For this setting, DANN and Deep Coral perform best, while DALN, DSAN and BNM suffer huge accuracy drop using ShuffleNet. However, when a batch size of 16 is used, the DALN algorithm on ShuffleNet exhibits the highest performance while using less GPU memory. In comparison, MobileNet has a higher GPU memory usage, comparable to that of Resnet50, which is not practical on edge devices or low-end devices. {Overall, DSAN and DALN show feasible improvements compared to without DA, highlighting their potential to adapt light-weight neural networks. However, when the number of samples in each batch is small (i.e., 4), the use of DA techniques does not take advantage of the class information (i.e., 4 samples cannot fully reflect the characteristics of the data), leading to negative adaptation.}


\begin{table}[h] \scriptsize
    \setlength{\tabcolsep}{1.1cm}
    \caption{{Top classification accuracy (\%) on COVID-19 datasets. The highest results are labeled with \textbf{bold} text.}}
    \begin{tabular}{c|cccc}
    \toprule
    Alg.& \multicolumn{3}{c}{Accuracy}&\\
    \midrule
  Batch size\,=\,16  & ShuffleNet & MobileNet & $\Delta$ & GMU \\
    \midrule
    DC &80.8 &80.6 &$+0.2$&\multirow{7}{*}{\rotatebox{90}{2.1, 5.0 GB}} \\
    DANN &78.7 &84.1 &$-5.4$& \\
    DSAN &{86.0} &{88.3} &$-2.3$& \\
    BNM &77.9 &80.4 &$-2.5$& \\
    DALN&\textbf{92.0} &\textbf{90.0} &$+2.0$ & \\
    DCAN&82.5 &84.9 &$-2.4$ &   \\
    w/o DA &83.6 &86.2 &$-2.6$& \\
    \midrule
  Batch size\,=\,8  &  &  &  &  \\
    \midrule
    DC &81.5 &85.8 &$-4.3$&\multirow{7}{*}{\rotatebox{90}{1.7, 3.1 GB}} \\
    DANN &79.0 &84.2 &$-5.2$& \\
    DSAN &\textbf{89.5} &89.3 &$+0.2$& \\
    BNM &78.6 &84.6 &$-6.0$& \\
    DALN&80.2 &84.3 &$-4.1$ & \\
    DCAN&80.5 &88.0 &$-7.5$ &   \\
    w/o DA &79.3 &\textbf{92.6} &$-13.3$& \\
    \midrule
  Batch size\,=\,4  &  &  &  &  \\
    \midrule
    DC &80.6 &90.9 &$-10.3$&\multirow{7}{*}{\rotatebox{90}{1.4, 2.2 GB}} \\
    DANN &79.6 &\textbf{91.4} &$-11.8$& \\
    DSAN &69.9 &85.7 &$-15.8$& \\
    BNM &58.4 &90.1 &$-31.7$& \\
    DALN&53.0 &78.5 &$-25.5$ & \\
    DCAN&81.0 &83.3 &$-2.3$ &   \\
    w/o DA &\textbf{87.3} &90.4 &$-3.1$& \\
    \bottomrule
    \end{tabular}
{GMU: GPU memory usage. $\Delta$ means the difference between ShuffleNet and MobileNet results.}
    \label{tab:COVID_Limit_Computation}
\end{table}


\textcolor{black}{\subsection{Out-of-distribution data for testing} \label{S:OOD}}

In the literature on DA, the test data used to evaluate the models' performance typically comes from the same dataset as the training data. Even though the source and target domains are different, this evaluation protocol may not assess the model's true capacity to generalize. To alleviate this issue, we conducted an empirical evaluation that measures the generalization capabilities of DA algorithms on out-of-distribution samples. Toward this goal, we employed PyTorch's data transformation tool to apply four distinct transformations on samples of the COVID-19 dataset: random flip (1), random inversion (2), Gaussian filtering (3), and RandomErasing (4). Algorithm \ref{Ag:2} details the augmentation procedure employed for this experiment, which uses Resnet50 as network backbone. Table \ref{tab:COVID_out_of} reports the classification accuracy of the methods tested in this scenario, and Figure \ref{F:COVID_Out_of_distribution} gives the accuracy of the methods over different training epochs. It can be seen that the inversion transform has the most notable impact on performance. Moreover, DSAN and DALN provide higher robustness for Gaussian blur and random flip, respectively. {These findings suggest that transformation on target domain can degrade the potential of the correlaton-based method (DC), while the MMD (DSAN, DCAN) and adversarial (DALN)-based approaches are more robust to these changes. }

\begin{algorithm} \scriptsize
\SetAlgoLined
    \PyComment{Define the data transformation pipeline based on four choices (each time only perform a single choice)} \\
    \PyCode{transformations = transforms.Compose([} \\
    \Indp
        \PyCode{transforms.Resize([256, 256]),} \\
        \PyCode{transforms.CenterCrop(224),} \\
        \PyCode{transforms.RandomHorizontalFlip(p=1),} \\
        \PyCode{transforms.RandomInvert(p=1),} \\
        \PyCode{transforms.GaussianBlur(kernel\_size=3),} \\
        \PyCode{transforms.ToTensor(),} \\
        \PyCode{transforms.RandomErasing(p=1)} \\
    \Indm
    \PyCode{])} \\
    \PyComment{For each data item in the dataset} \\
    \PyCode{for data in Data:} \\
    \Indp
        \PyComment{Apply the transformations to the data} \\
        \PyCode{transformed\_data = transformations(data)} \\
        \PyComment{Append the transformed data to the list of transformed data} \\
        \PyCode{Transformed\_data.append(transformed\_data)} \\
    \Indm
\caption{Data transformation procedure for out-of-distribution scenario}
\label{Ag:2}
\end{algorithm}

\begin{table}[h] \scriptsize
    \setlength{\tabcolsep}{0.75cm}
    \caption{{Summary of the highest classification accuracy (\%) on COVID-19 dataset with out-of distribution case with Resnet50 as backbone. The best outcomes are highlighted with \textbf{bold} text.}}
    \begin{tabular}{c|cccc}
    \toprule
    Alg.& \multicolumn{4}{c}{Accuracy}\\
    \midrule
    & Random Flip & Random Invert & Gaussian Blur & Random Erasing \\
    \midrule
    DC &80.7 &54.6 &80.9&62.1 \\
    DANN &83.5 &66.7 &79.2&59.5 \\
    DSAN &{88.3} &{69.9} &\textbf{86.9}&{76.1} \\
    BNM &80.2 &61.8 &80.4&61.5 \\
    DALN&\textbf{94.5} &64.6 &86.2 &\textbf{83.4}  \\
    DCAN&80.7 &\textbf{70.2} &81.5 &{79.8}   \\
    w/o DA &81.1 &62.6 &81.0&68.7  \\
    \bottomrule
    \end{tabular}
    \label{tab:COVID_out_of}
\end{table}

\begin{figure}
     \centering
    \includegraphics[width = 0.97 \textwidth]{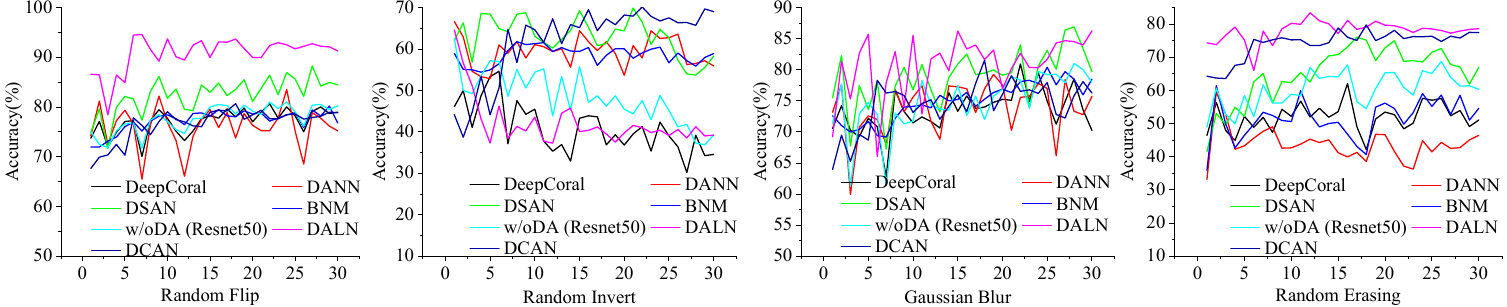}
    \caption{{Test classification accuracy on COVID-19 dataset for out-of-distribution scenario. The columns, from left to right, represent Resnet34, Resnet50, and Densenet121, respectively. The x-axis represents the number of training epochs.}}
    \label{F:COVID_Out_of_distribution}
\end{figure}

\textcolor{black}{\subsection{Impact of limited training samples}}
In the field of healthcare, there is often a lack of available data for certain diseases, resulting in an insufficient amount of training samples. To evaluate DA algorithms in this challenging scenario, we chose the COVID-19 dataset and randomly selected 100 samples from each label to create a new training set. We then tested the trained models using the standard test set of the COVID-19 dataset. Details about this new dataset, named LCOVID-19, can be found in Table \ref{tab:Dataset_Details}. For this experiment, we considered the Resnet34, Resnet50, and Densenet121 architectures as backbone and used the parameters in Table \ref{T:Details_Alg} for training the models. Table \ref{tab:LCOVID} reports the classification results of these models, while Figure \ref{F:COVID_Limited} presents the models accuracy at each training epoch. We can see that most DA algorithms lead to improved performance when using Resnet50, with DALN displaying the most notable gains. {When using the smaller training sets, for Resnet34, DSAN (50.5\%), DC (45.8\%), BNM (48.4\%) and DANN (48.6\%) perform lower than the CNN baseline without DA (52.1\%). This may be due to the lower learning capacity of Resnet34 compared to Resnet50, which makes DA algorithms such as DC not learn the knowledge of the correlation domain related to the task, which ultimately leads to deterioration in overall performance.}

\begin{table}[h] \scriptsize
    \setlength{\tabcolsep}{1.33cm}
    \caption{{Summary of the highest classification accuracy (\%) on LCOVID-19 datasets. The best outcomes are highlighted with \textbf{bold} text.}}
    \begin{tabular}{c|ccc}
    \toprule
    Alg.& \multicolumn{3}{c}{Accuracy}\\
    \midrule
    & Resnet34 & Resnet50 & Densenet121 \\
    \midrule
    DC &45.8 &48.5 &45.6 \\
    DANN &48.6 &49.9 &58.5 \\
    DSAN &50.5 &{58.3} &{58.8} \\
    BNM &48.4 &52.9 &43.6 \\
    DALN&\textbf{60.3} &\textbf{64.9} &\textbf{62.3}  \\
    DCAN &59.6 &59.7 &59.1   \\
    w/o DA &{52.1} &50.1 &44.3  \\
    \bottomrule
    \end{tabular}
    \label{tab:LCOVID}
\end{table}

\begin{figure}[htp]
     \centering
    \includegraphics[width = 0.98 \textwidth]{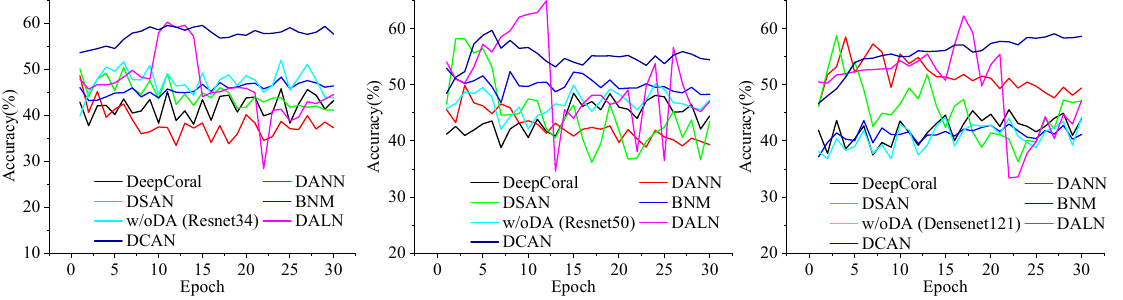}
    \caption{{Test classification accuracy using six DA techniques on LCOVID-19 dataset. The columns, from left to right, represent Resnet34, Resnet50, and Densenet121, respectively.}}
    \label{F:COVID_Limited}
\end{figure}

\textcolor{black}{\subsection{Analysis of $\mathcal{A}$-distance} }

\textcolor{black}{The $\mathcal{A}$-distance is a metric commonly used to estimate the similarity between the two distributions \cite{ben2006analysis}. Concretely, the A-distance between two distributions $\mathcal{P}$ and $\mathcal{P}'$ is defined as 
\begin{equation}
d_\mathcal{A}(\mathcal{P}, \mathcal{P}') \, = \, 2 \sup_{A\in \mathcal{A}} |\mathcal{P}(A) - \mathcal{P}'(A) |, 
\end{equation}
where $\mathcal{A}$ is a collection of measurable sets over the same space as $\mathcal{P}$ and $\mathcal{P}'$. In practice, the $\mathcal{A}$-distance is estimated using the smallest error of a classifier $h \in \mathcal{H}$ discriminating between points sampled from the two distributions:
\begin{equation}
d_\mathcal{A}(\mathcal{P}, \mathcal{P}') \, \approx \, 2 \Big(1 \, - \, 2 \min_{h\in \mathcal{H}} \, \mathrm{err}(h)\Big).
\end{equation}
Figure \ref{F:A_distance} shows the $\mathcal{A}$-distance measured for DA algorithms on the Kidney cancer, Skin cancer, Cervical cancer, and Retinal disease datasets, where a smaller distance indicates a better domain alignment. We see that DSAN achieves notably better results than other algorithms, as illustrated by the decreasing distance between the source and target domains. {Deep Coral, DANN, DCAN and BNM exhibit similar results, with $\mathcal{A}$-distance values smaller than the baseline without DA.} The effectiveness of DSAN, which explicitly enforces the alignment of domains, is apparent in this scenario.} {These findings suggest that LMMD captures more dominant task-related features, thereby provides lower $\mathcal{A}$-distance.}

\begin{figure}
     \centering
    \includegraphics[width = 0.97 \textwidth]{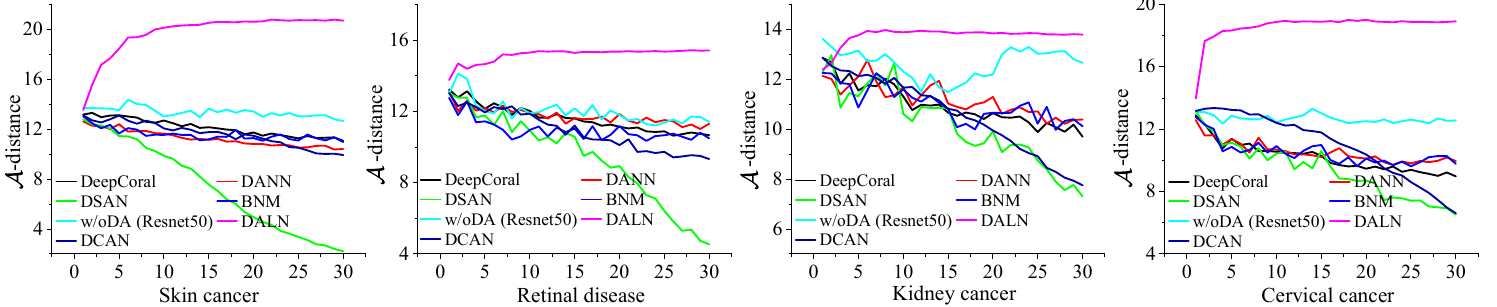}
    \caption{{$\mathcal{A}$-distance for the DA techniques using four datasets (i.e., skin cancer, retinal disease, kidney and cervical cancer). The Y-axis represents the $\mathcal{A}$-distance value, whereas the X-axis represents the training epochs.}}
    \label{F:A_distance}
\end{figure}

\textcolor{black}{\subsection{Dynamic data stream domain adaptation}\label{sec:dynamicDA}}
So far, our experiments have considered a fixed distribution for the source and target domains. However, this scenario may be too simplistic for real-world situations where changes can occur over time. In practice, we also want a model that can dynamically adapt to domain shifts in a stream of data. To simulate this dynamic data stream scenario of DA, we performed two experiments related to computer vision and medical imaging. Specifically, we used the Adaptiope dataset, considering domain P as the source domain and domain R as the target domain. We also used the COVID-19 dataset, where the training set was considered as the source domain and the testing set as the target domain. Since domain P of the Adaptiope dataset is smaller than the COVID-19 training set, we divided the former into 30 non-overlapping parts and then trained the model for 30 epochs, each one using a different (single) portion of the data. The same test set was employed for evaluation throughout the training process. We used a similar strategy for the COVID-19 dataset, except that the training set was divided into 100 non-overlapping parts, and we used 100 epochs for training. {Table \ref{T:dynamic_Data} reports the highest classification accuracy of the DA algorithms, and Figure \ref{F:Dynamic_DA} shows their testing accuracy measured during training epochs.} The accuracy values of the algorithms tested on the Adaptiope dataset are nearly identical, except DALN shows an accuracy drop. This suggests that for large scale dynamic natural datasets, the performance of DALN is limited as it fails to capture the useful domain information. However, for the COVID-19 dataset, DALN exhibits higher performance over other DA methods, highlighting the potential of DALN for COVID-19 data adaptation. Furthermore, DCAN and DSAN indicate better performance compared to Deep Coral, DANN, BNM and Resnet34 alone. {These results suggest that MMD (DSAN, DCAN) and adversarial based (DALN) domain alignment can effectively manage domain shifts occurring in data streams related to medical images, while DALN can not adapt the dynamic features in natural images.}

\begin{table}[h]\scriptsize
    \setlength{\tabcolsep}{2.27cm}
    \caption{{Classification accuracy (\%) using dynamic data stream environment. The best accuracies are highlighted with \textbf{bold} text.}}
    \begin{tabular}{c|cc}
    \toprule
     Adaptiope & Accuracy & $\Delta$  \\
     \midrule
     DC &65.0 &-3.4   \\
     DANN &66.0 &-3.6   \\ 
     DSAN &66.4 &-5.8   \\ 
     BNM &\textbf{69.0} &-5.9   \\
     DALN&55.5 &-23.7 \\
     DCAN &67.7 &-6.8  \\
     w/o DA &63.4 &-3.9   \\
     \midrule
     COVID-19 & & \\
     \midrule
     DC &55.3 &-25.4   \\
     DANN &57.6 &-22.8   \\ 
     DSAN &{75.4} &-14.4   \\ 
     BNM &70.4 &-14.1   \\ 
     DALN&\textbf{93.5} &-1.0 \\
     DCAN &80.7 &-8.2  \\
     w/o DA &68.7 &-21.7   \\
     \bottomrule
    \end{tabular}
   {$\Delta$ represents the accuracy differences compared to Table \ref{tab:adaptiope} and Table \ref{tab:summary}. }
    \label{T:dynamic_Data}
\end{table}

\begin{figure}[htp]
     \centering
    \includegraphics[width = 0.99 \textwidth]{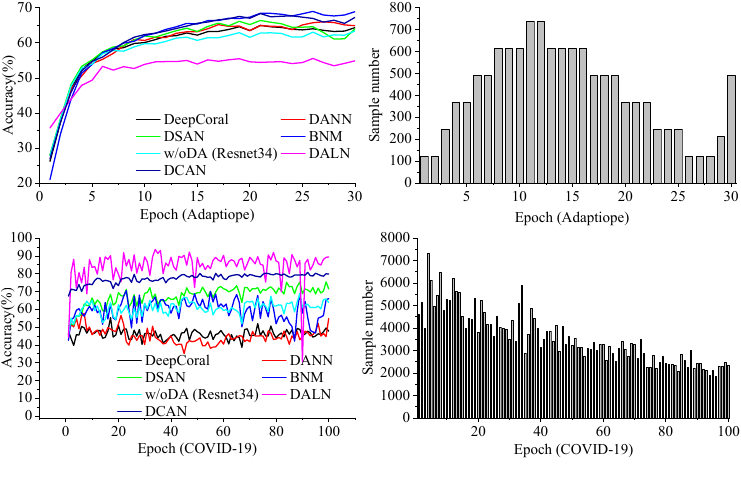}
    \caption{{The left column represents the classification accuracy achieved with six DA techniques on the Adaptiope and COVID-19 datasets, with the x-axis indicating the training epochs. The right column shows the sample number in each data stream.}}
    \label{F:Dynamic_DA}
\end{figure}

\textcolor{black}{\subsection{Interpretability of the model}\label{sec:interpret}} 

The interpretability of an AI model is crucial for allowing clinicians and patients to understand its predictions. Having a model that accurately predicts outcomes but does not prioritize the disease itself can be risky. To assess the interpretability of our DA approaches, we used Grad-CAM \cite{selvaraju2017grad} for visualizing important regions in images of a specific category within the COVID-19 and skin disease datasets. Specifically, we used the fourth convolutional layer of the Resnet34 model as a feature map for visualization. Figure \ref{F:XAI} shows the heatmaps obtained using Grad-CAMs on the different algorithms. For the COVID-19 dataset (first row), while most DA algorithms and the baseline without DA locate the main abnormal regions in the image, DSAN and BNM provide more interpretable results as the high-activation values in their heatmap are restricted to specific regions. 
In comparison, the heatmaps of other algorithms and the baseline have more spread activations, making their interpretation difficult. Although the baseline without DA achieves a high classification accuracy of 90.4\%, its trustworthiness is questionable due to its pronounced focus on bone tissues, which lacks clinical relevance. Similar results are observed for the skin lesion image in the second row. In this case, the Grad-CAM obtained with Deep Coral properly highlights the abnormal area. {These findings suggest that introducing DA does not guarantee meaningful interpretability. Instead, specific techniques, such as attention modules, can be employed to shift the model attention to the relevant regions.}

\begin{figure}[htp]
     \centering
    \includegraphics[width = 0.97 \textwidth]{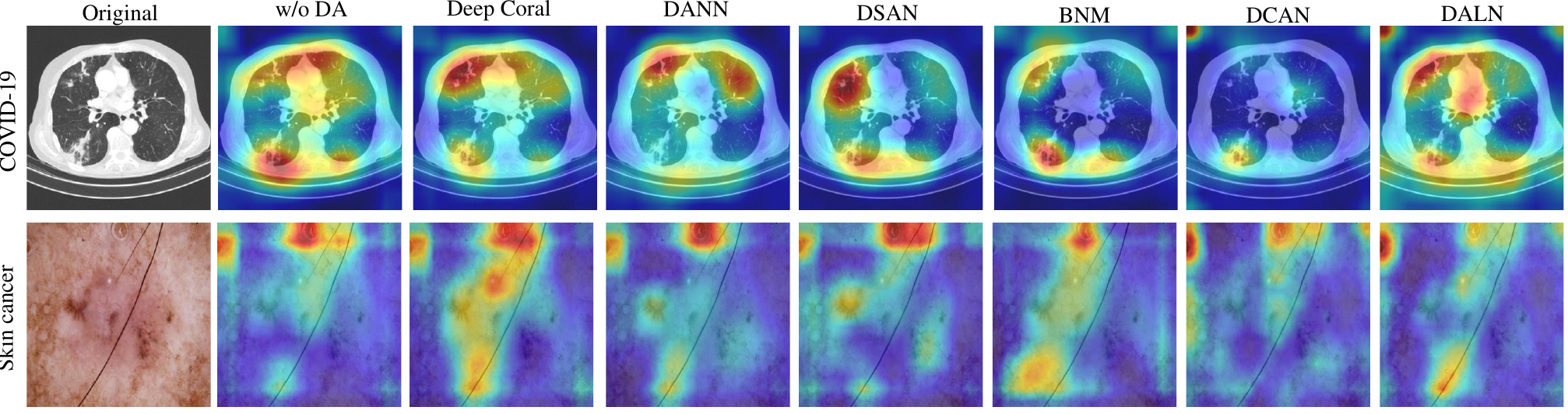}
    \caption{{Example of heatmap visualization using Grad-CAM. The model's attention is more focused on its current location when the color is deeper, such as red.}}
    \label{F:XAI}
\end{figure}

\section{Discussion}\label{S:5}

Our experiments on medical images revealed that, while the tested DA algorithms demonstrate exceptional performance on small or high-quality datasets (e.g., Kidney cancer or cross-dataset), their performance dropped significantly for highly-challenging datasets such as ChestXray8. This emphasizes the need for developing more robust DA methods. Experimental results using multisource skin cancer datasets indicate that for large-scale datasets with class imbalance, the use of DA may lead to reduced adaptation, highlighting the necessity of choosing the most suitable adaptation method. {We argue that MMD and its variants are better suited for class imbalanced data sets (i.e., they provide higher test metrics). In addition, future work can leverage the diffusion model to help the DA model learn rare class features by generating high-quality samples.} The results obtained for different network backbones (Resnet34, Resnet50 and Densenet121) also underscore the importance of not relying solely on more powerful neural networks to address the problem of distribution shifts in the data. 

{Similarly, the experimental results as reported in Table \ref{tab:MultiImageNet} using MultiImageNet dataset demonstrate that for large-scale natural images with huge texture shifts, the use of DA exhibits limited feature adaptation ability. Figure \ref{fig:TextureShiftsss} shows a simple example of texture and style shifts in MultiImageNet with an orange circle denoting the task specific class. For example, ImageNet-R shows a large shift compared to ImageNet-M. {Intuitively, direct adapting all features is not practical since only small parts related to the specific class, this poses a challenge for DA. DA can then be used with the attention methodology \cite{10413589} to improve the performance of the model by shifting its attention to tasks-related regions and ignoring unrelated parts. }

\begin{figure}
    \centering
    \includegraphics[width = 0.97\textwidth]{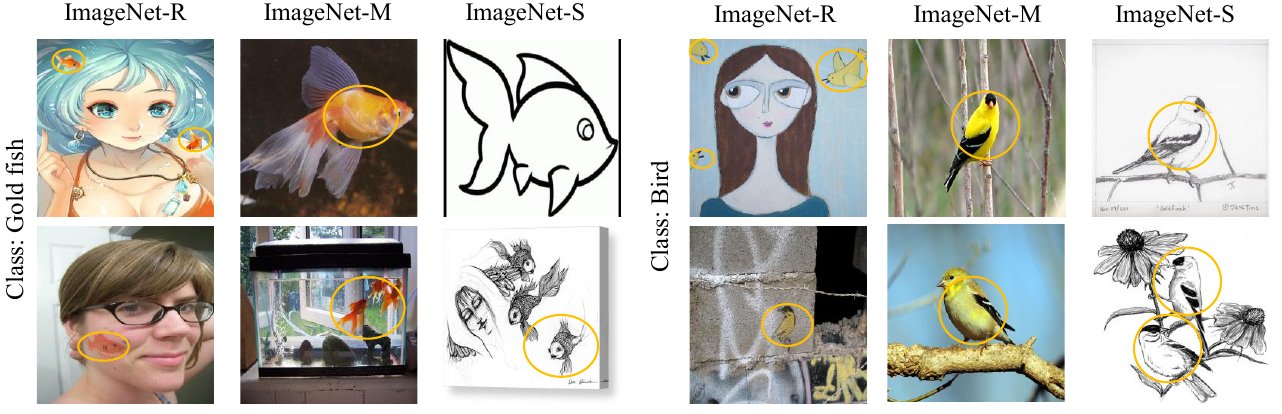}
    \caption{{Example of texture and style shifts in MultiImageNet dataset. Orange circle indicates the specific class of this image. ImageNet-R shows wide shifts where only small parts represent the class specific object.}}
    \label{fig:TextureShiftsss}
\end{figure}

{Furthermore, the implementation of stringent data privacy protection laws and regulations introduces  substantial challenges in training machine learning models with medical data. Since DA requires access to source or target data, this poses a domain-specific challenge because the source data may not be available.} FL, an emerging framework for distributed learning, offers a potential solution to this problem by enabling the training of multiple local models through the federation of diverse datasets, without requiring direct access to the source data \cite{10288131}. 
Nevertheless, FL approaches for medical imaging still have to deal with the issue of heterogeneity in the data from different sites \cite{10288131}. This leads to a potential avenue of research that is the combination of FL and DA techniques. {Future work can explore FL with DA without requiring a source dataset (e.g., use a GAN or a diffusion model to simulate data characteristics), which is more practical in the medical domain.} 

{Similarly, the question of the interpretability of machine learning models, including those for DA, is the subject of ongoing research. An interpretable DA model can improve the reliability of predictions and trust from patients.} {The t-SNE plots obtained for DSAN on the MultiImageNet dataset (Fig.~\ref{F:TNSE_Medical_All}), illustrate the challenging nature of interpreting the results of the DA algorithms. However, the use of DA can demonstrate higher interpretability compared to CNN alone in skin cancer dataset.} Moreover, while DANN, Deep Coral and BNM have an accuracy comparable to DSAN for the retinal disease dataset (Table~\ref{tab:summary}), they exhibit negligible reductions in $\mathcal{A}$-distance compared to that algorithm (Fig.~\ref{F:A_distance}). Further analyses could be performed to better understand how the former algorithms reduce discrepancies in the data. Future studies should also prioritize the development of interpretable methods in the field of DA (e.g., XAI \cite{selvaraju2017grad}), with the aim of enhancing the usability of such tools for medical professionals and patients alike. 

Most of our experiments used large sets of labeled data for training models. However, in medical imaging, it is more likely to encounter a scenario where only a limited number of samples, ranging from a few to a dozen, are labeled. {The lack of labeled data can limit the feature representation ability of the neural networks, thus decreasing the feature adaptation reliability (i.e., mis-alignment). Future work can explore the use of few-shot DA techniques \cite{sun2025adversarial}.} We can also take advantage of the limited data available by setting up an upstream task, such as randomly rotating the images and categorizing the degrees, to improve the feature extraction capabilities of the DA model. This can then be transferred to a downstream task, such as classifying the disease classes, which may result in an improved model performance. 


\begin{table}[h]
    \renewcommand{\arraystretch}{0.9}
    \caption{A summary of strengths and weaknesses of these DA techniques.}
    \setlength{\tabcolsep}{0.5cm}
    \begin{tabular}{c m{6cm} m {5.5cm}}
    \toprule
     Alg.& Strengths & Weaknesses  \\
     \midrule
     Deep Coral & Robustness for different optimizers; High performance with small batch size with light neural networks & Less competitive when using large-scale datasets; Poor adaptability to out-of-distribution data.  \\
     \midrule
     DANN & More useful for dealing with large-scale labels; Excellent performance when using a small batch size with the light neural network. & Less effective for large-scale medical datasets. \\
     \midrule
     DSAN & Very effective in mitigating data distribution discrepancies; The interpretability of the model is decent; Highly capable of handling dynamic data  & Performance is dependent on batch size. \\
    \midrule
    DCAN&{Solid for dynamic COVID-19 data, effective in minimizing data distribution discrepancies}&{Less effective for cross-dataset; unstable under small batch size for medical data} \\
    \midrule
    DALN&{Useful for Medical data, high interpretability, robust for out-of-distribution medical data}&{Easy to overfit with small batch size, not robust for dynamic natural data} \\
     
     \midrule
     BNM & Robustness for dynamic data distribution  & Likely to overfit using small batch size; The model is slightly less interpretable; Unstable for small batch size.  \\
     \bottomrule
    \end{tabular}
    \label{tab:strengths_weaknesses}
\end{table}

In practice, it is possible for the quality of the data to be poor, or the labels associated with the data to be compromised (e.g., Section \ref{S:OOD}). {Since the DA model requires correct source data information, such a situation leads to a decrease in the model performance. These situations pose a challenge called noisy DA}. Developing robust domain-adaptive models is a formidable task, as it involves addressing the potential negative impact of data quality. {Future work can explore the use of FMs as the backbone, as their feature extraction ability is robust to noise.} In addition, many existing algorithms for DA suppose a static distribution \cite{zhu2020deep} within a dataset. However, this claim rarely holds in practical situations, where data tend to be dynamic in nature. Our experiments on dynamic data streams (Section \ref{sec:dynamicDA}) showed that DA algorithms suffer a substantial drop in performance when dividing the COVID-19 dataset into non-overlapping subsets and training the model sequentially on these subsets. {Future work can explore DA that can continuously adapt to new data, without having access to previously seen datasets (e.g., test time adaptation that adapts the model with continuous test data).}



Based on our XAI experiments (Section \ref{sec:interpret}), we observed that DA algorithms can be influenced by noise (e.g., scars on skin). In the COVID-19 dataset, we found that the use of DA could enhance the classifier's ability to prioritize and detect abnormal regions. However, when analyzing the Grad-CAM heatmaps obtained for the tested algorithms (Fig. \ref{F:XAI}), we noticed differences in their interpretability. In particular, DSAN gave Grad-CAMs which focused on more relevant regions than those of other algorithms. {Future work can investigate the development of DA methods that are more interpretable to address the challenges of interpretability.}

The strengths and weaknesses of the DA techniques considered in this work are summarized in Table \ref{tab:strengths_weaknesses}. Despite conducting numerous experiments, our study has several limitations that could be addressed in future work. For example, the impact of additional training hyperparameters, such as learning rate and weight decay, could be investigated. Moreover, we only evaluated cross-entropy as a training loss for classification and omitted popular alternatives like focal loss, which were shown to produce better-calibrated networks \cite{mukhoti2020calibrating}. In the future, we will conduct a more thorough analysis taking into account these various aspects.


\section{Conclusion}\label{S:6}
The work in \cite{recht2018cifar} explored the question of whether classifiers trained on a given dataset generalize to other examples from the same dataset. This work underlined the problem that the popular benchmark test set is often used to select the best model during training, thereby overestimating the accuracy of the models. Similarly, the accuracy reported for the best-performing DA methods may not be reliable, since it was measured on test sets that have been around for many years (e.g., Office31, CLEF, etc.). In this study, we used 13 natural and medical imaging data sets to evaluate the performance of six common / new DA algorithms: Deep Coral, DANN, DSAN, DALN, DCAN, and BNM.} The results of our experiments demonstrated that the use of DA can greatly improve image classification performance on computer vision data sets. However, for most medical imaging datasets, these algorithms provided a limited improvement over baseline without DA. However, we found that DA algorithms could improve the accuracy of classifiers in situations with limited training data or with out-of-distribution cases. This finding is important because it can guide the development of robust AI models for medical imaging. Furthermore, our investigation underlined the challenge of learning robust models from dynamic data streams, and showed that DA techniques can alleviate the impact of distribution shifts in this type of data. This benefit was particularly evident for the COVID-19 dataset, where the use of DA substantially improved the classification accuracy. These results can provide guidelines for the future development of DA models in the medical field. 

\section*{Acknowledgments}
This research was funded by the National Natural Science Foundation of China grant number 82260360, the Innovation Project of GUET Graduate Education 2025YCXS244, and the Guangxi Science and Technology Base and Talent Project (2022AC18004, 2022AC21040). 

{
\bibliographystyle{unsrt}
\bibliography{cas-refs}
}

\end{document}